\documentclass[10pt,twocolumn,letterpaper]{article}

\usepackage{iccv}              % To produce the CAMERA-READY version
\definecolor{iccvblue}{rgb}{0.21,0.49,0.74}
\usepackage[pagebackref,breaklinks,colorlinks,allcolors=iccvblue]{hyperref}
\usepackage{color, colortbl}

\definecolor{LightBlue}{RGB}{212, 250, 252} 
\definecolor{LightGreen}{RGB}{217, 250, 226} 
\newcommand{\loss}[1]{\mathcal{L}_{#1}}

\def\paperID{8} % *** Enter the Paper ID here
\def\confName{ICCV}
\def\confYear{2025}

\usepackage{tikz}
\newcommand\copyrighttext{%
  \footnotesize \textcopyright 2025 IEEE.  Personal use of this material is permitted.  Permission from IEEE must be obtained for all other uses, in any current or future media, including reprinting/republishing this material for advertising or promotional purposes, creating new collective works, for resale or redistribution to servers or lists, or reuse of any copyrighted component of this work in other works.}
\newcommand\copyrightnotice{%
\begin{tikzpicture}[remember picture,overlay]
\node[anchor=south,yshift=10pt] at (current page.south) {\fbox{\parbox{\dimexpr\textwidth-\fboxsep-\fboxrule\relax}{\copyrighttext}}};
\end{tikzpicture}%
}

\title{Robust Experts: the Effect of Adversarial Training on CNNs with Sparse Mixture-of-Experts Layers}

\author{Svetlana Pavlitska$^{1,2}$, Haixi Fan$^{2}$, Konstantin Ditschuneit$^{1}$, J. Marius Zöllner$^{1,2}$\\
\textit{$^{1}$ Karlsruhe Institute of Technology (KIT), Germany}\\
\textit{$^{2}$ FZI Research Center for Information Technology, Germany} \\
{\tt\small pavlitska@fzi.de}\\
}

\begin{document}
\maketitle
\copyrightnotice
\thispagestyle{empty}
\pagestyle{empty}

\begin{abstract}
Robustifying convolutional neural networks (CNNs) against adversarial attacks remains challenging and often requires resource-intensive countermeasures. We explore the use of sparse mixture-of-experts (MoE) layers to improve robustness by replacing selected residual blocks or convolutional layers, thereby increasing model capacity without additional inference cost. On ResNet architectures trained on CIFAR-100, we find that inserting a single MoE layer in the deeper stages leads to consistent improvements in robustness under PGD and AutoPGD attacks when combined with adversarial training. Furthermore, we discover that when switch loss is used for balancing, it causes routing to collapse onto a small set of overused experts, thereby concentrating adversarial training on these paths and inadvertently making them more robust. As a result, some individual experts outperform the gated MoE model in robustness, suggesting that robust subpaths emerge through specialization. Our code is available at \url{https://github.com/KASTEL-MobilityLab/robust-sparse-moes}.
\end{abstract}

\section{Introduction}
\label{sec:intro}

Deep neural networks achieve state-of-the-art performance in natural language processing and computer vision but still suffer from inherent limitations, including adversarial brittleness~\cite{goodfellow2014explaining,szegedy2013intriguing,madry2017towards}. % Adversarial defenses attempt to robustify neural networks artificially, but robustly solving a task fundamentally increases its difficulty. For example, classifying images is much more difficult if an adversarial attacker can search for weaknesses and uncertainties in the decision process. It comes thus as no surprise that more capable networks, while requiring significantly more computing resources, can generally solve tasks more robustly. However, simply scaling model sizes is not always an option and is quickly restricted by technical and financial factors.
A sparse mixture of experts (MoE) is a type of neural network architecture that selectively activates only a subset of parameters for each input instead of using the entire model~\cite{shazeer2017outrageously,lepikhin2021gshard,du2022glam}. This approach allows extremely large models, reaching trillions of parameters, to be trained efficiently, saving compute resources since only a fraction of the model is used for each input. Furthermore, experts can learn different data subdomains, improving performance and interpretability. While the concept has originated in the field of natural language processing and focused on transformers, sparse MoE layers for CNNs were also proposed, with experts as filters~\cite{wang2019deep,zhang2023robust} or layers~\cite{pavlitska2023sparsely}.

\begin{figure}[t]
\centering
    \includegraphics[width=0.9\columnwidth]{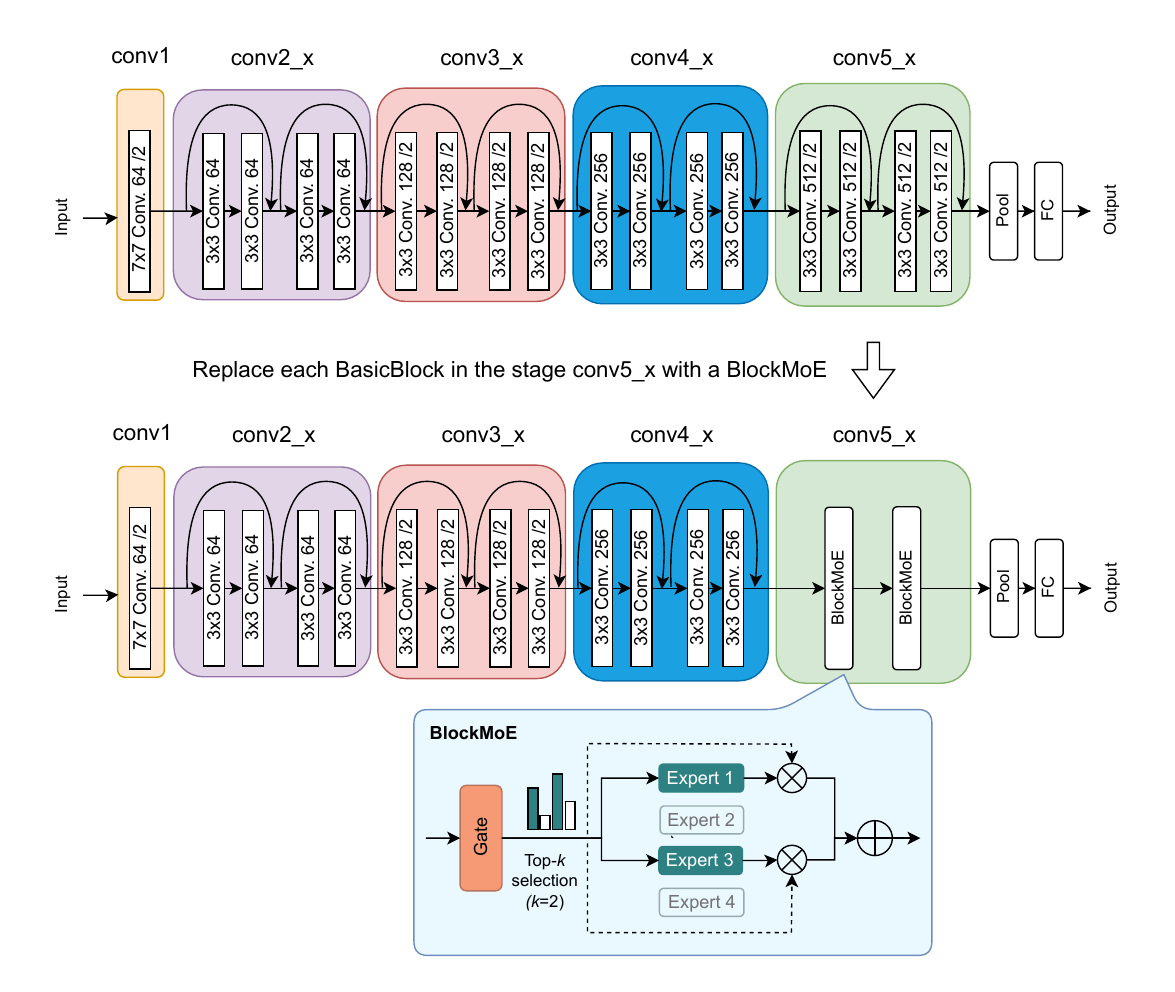}
    \caption{We analyze the impact of embedding sparse MoE layers in CNNs at the level of BasicBlocks or convolutional layers on the adversarial robustness. Here, each BasicBlock in the $conv\_5x$  stage in a \texttt{ResNet-18} is replaced with a BlockMoE layer with four experts and a gate with top-$2$ routing. Each expert has the architecture of a replaced BasicBlock.}
    \label{fig:concept}
\end{figure}

Apart from boosting model accuracy, sparse MoE layers can lead to better adversarial robustness by promoting expert specialization and reducing shared vulnerabilities through input-dependent, selective computation paths.
Previous works have demonstrated the potential of the sparse MoE layers for increasing adversarial robustness, focusing either on transformers~\cite{puigcerver2022onthe} or MoE layers in CNNs with filters as experts~\cite{zhang2023robust}. We consider MoE layers at the level of coarse-grained structures like residual blocks (see Figure~\ref{fig:concept}) to enhance robustness by allowing each expert to learn more semantically meaningful and independent representations, reducing interference between experts and improving gradient isolation during adversarial training.

%Modifications to existing CNN architectures based on sparse MoE would require little hardware resources but may provide increased robustness against adversaries through more trainable parameters. Positive results would be a considerable step towards deploying more robust neural networks in the real world.

\newpage
\section{Related Work}

\subsection{Mixture of Experts}

The idea of dynamic routing in a Mixture of Experts (MoE)  dates back to the early work of Jacobs et al.~\cite{jacobs1991adaptive}, their original MoE model involved a gating network that directed inputs to different subnetworks based on the input's characteristics. This approach aimed to divide learning tasks across experts. However, it was computationally expensive because all experts were active for every input.

Eigen et al.~\cite{eigen_learning_2013} developed the idea of using MoEs as sub-components of models with individually learned gating networks. This allows for sharing the remaining parts of the architecture and enables multiple MoE layers within a single architecture.  Shazeer et al.~\cite{shazeer2017outrageously} first explored sparse activations using LSTM networks. This led to renewed interest in MoE models, especially for large-scale tasks. Architectures like the Switch Transformer\cite{fedus2022switch} and GLaM~\cite{du2022glam} showed that it's possible to train massive models efficiently by using only a subset of experts at a time. Although sparse MoEs are mostly used in language modeling, the general idea has also been explored in other areas like vision and multimodal learning. Recently, DeepSeek-MoE~\cite{dai2024deepseekmoe} demonstrated the continued scalability of MoE-based models, achieving strong performance with reduced inference cost.

In computer vision, MoEs have been integrated with vision transformers through the V-MoE framework of Riquelme et al.~\cite{riquelme2021scaling}, while CNNs have seen similar integration beginning with DeepMoE~\cite{wang2019deep}, which replaced individual convolutional layers with sparsely-gated modules. Pavlitska et al.~\cite{pavlitska2023sparsely} advanced this line of work by substituting entire residual blocks with MoE layers, with the aim of enhancing interpretability and structural modularity. Parallel lines of inquiry have also explored model-level MoE mechanisms in various computer vision settings~\cite{mees2016choosing,valada2016convoluted,ma2018modeling,pavlitskaya2020using,xu2025limoe,zhang2019learning}, reaffirming the versatility of sparse expert architectures in structured visual domains.

\subsection{Expert Balancing and Routing Collapse}
\label{subsec:expert_balancing}
%\todo{Possibly elaborate more on (unused/unrelated) types of expert balancing}

Routing collapse~\cite{shazeer2017outrageously} occurs in sparse MoEs when the router consistently assigns most inputs to a small subset of experts, leaving others underutilized. Combining random weight initialization with random batches of samples leads to experts receiving different importance scores even before the first gradient descent step is applied. This can lead to gradient signals of different magnitudes propagating through individual experts. Some experts often do not receive any gradient signal at all if the gating network initially does not assign them a large enough probability for any input. The greedy nature of stochastic mini-batch gradient descent reinforces the effect by further improving the experts that initially receive high importance. This effect can also be called the \textit{dying expert} phenomenon~\cite{pavlitska2023sparsely}. 

This imbalance reduces model efficiency, leads to gradient starvation for underused experts, and prevents the model from leveraging its full capacity. To mitigate this, several techniques are used for expert balancing. The load balancing losses, such as the importance loss~\cite{shazeer2017outrageously} or the switch loss,\cite{fedus2022switch} encourage the router to distribute tokens more evenly by penalizing deviations from expected expert utilization. Temperature scaling in the router’s softmax function~\cite{nguyen2024is} smooths probability distributions, preventing extreme expert favoritism. Noisy routing~\cite{chowdhury2023patch,do2023hyperrouter}, where small random perturbations are added to router logits during training, encourages exploration and better load distribution. Additionally, capacity constraints~\cite{wang2024remoe} can limit the number of tokens any expert can process, forcing overflow tokens to be reassigned.

\subsection{Adversarial Robustness of Sparse MoEs}

Sparse MoE layers can enhance adversarial robustness by leveraging their architecture, which selectively activates subsets of specialized experts for each input. This selective routing can contribute to improved robustness through several mechanisms. Puigcerver et al.~\cite{puigcerver2022onthe} state that sparse MoE models can achieve a smaller Lipschitz constant compared to dense models, indicating reduced sensitivity to input perturbations. This reduction is attributed to the model's ability to route inputs to specific experts, effectively partitioning the input space and limiting the impact of adversarial perturbations. Puigcerver et al. were the first to explore the adversarial robustness of the models with embedded MoE layers. %They compared V-MoE~\cite{riquelme2021scaling} to ViT models on JFT-300M and ImageNet data under PGD attacks. V-MoE was shown to perform similarly to ViT. 
Furthermore, introducing sparsity into neural networks has been associated with improved adversarial robustness~\cite{guo2018sparse}. Sparse MoE architectures, by design, activate only a subset of experts, promoting sparsity and potentially enhancing the model's resistance to adversarial attacks. 

Several works have empirically explored the adversarial robustness of models with sparse MoEs. Zhang et al. have investigated the adversarial robustness of MoEs at the CNN filter level~\cite{zhang2023robust}. First, they showed that although models with sparse MoEs have a larger capacity than their dense counterparts, they behave similarly under adversarial training. Furthermore, they proposed a method to robustify routers and pathways jointly based on analyzing their impact. The reliability of models with sparse MoEs regarding hallucination, adversarial robustness, and robustness to out-of-distribution inputs was studied by Chen et al.~\cite{chen2024moebbench} with a focus on large language models. The authors stress the importance of the hyperparameter selection, training, and inference processes in the robustness of sparse models. They, however, conclude that models with MoE layers can exhibit better robustness compared to the dense baselines. Finally, Kada et al.~\cite{kada2025robustifying} describe a method to robustify routing in MoEs in visual transformers for image classification. 

\newpage
\section{Method}
\label{chapter:moe_architectures}

We propose two types of MoE layers: \textbf{BlockMoE} replacing a residual block and \textbf{ConvMoE} replacing a convolutional layer. 

\subsection{MoE Definition}

%An MoE uses a set of experts with individually learnable weights. A single input $x$ is distributed to a set of experts $\{E_i\mid i\in 1\ldots n\}$, and their outputs are weighted and summed up. A trained gating network $G$ computes \textit{importance scores} $P(x, i)$ for each input $x$. The output of the MoE is then the weighted sum: $\moe(x) = \sum_{i=1}^{n}P(x, i)\cdot E_i(x)$.
% \begin{equation}
%     \moe(x) = \sum_{i=1}^{n}P(x,i)\cdot E_i(x).
% \end{equation}

An MoE consists of \( N \) experts \( E_1, \dots, E_N \). For an input \( x \), each expert \( E_i \) produces \( e_i(x) \), and the gating network outputs scores \( G(x) = [g_1(x), \dots, g_N(x)] \). In sparse top-\( k \) routing~\cite{shazeer2017outrageously}, only the \( k \) experts with the highest scores are selected. The MoE output is:
\[
F_{\text{MoE}}(x) = \sum_{i=1}^{N} g_i(x) \cdot e_i(x),
\]
where \( g_i(x) = 0 \) for unselected experts. Expert utilization over a batch \( X \) is measured by the importance vector:
\[
I(X) = \sum_{x \in X} G(x), \quad I_i(X) = \sum_{x \in X} g_i(x).
\]
\subsection{MoE Layer Architecture}

%Replacing the last fully connected layer of a classification architecture is an easy way to provide the network with additional capacity to learn separated and more complex decision surfaces. This approach has been used to integrate MoE layers into CNNs before. Still, fully-connected layers are used in only a subset of CNNs, and replacing the fully-connected layer prevents us from gaining knowledge that is transferable to CNNs in general. Prior work~\cite{rajbhandari2022deepspeed} suggests that MoE layers should be located toward the end of deep neural networks.

Previous works embedding MoE layers considered replacing both fine-grained (e.g., convolutional filters~\cite{zhang2023robust} and convolutional layers~\cite{wang2019deep}) and coarse-grained (e.g., residual blocks~\cite{pavlitska2023sparsely}) DNN structures with an MoE layer. While focusing on transformers, the authors of the DeepSpeed-MoE~\cite{rajbhandari2022deepspeed} argue that placing MoEs at larger structural levels improves training stability and efficiency. We choose to focus on larger structural units, such as layers and residual blocks, with a higher capacity and semantic meaning. Compared to fine-grained filter-level MoEs, block-level MoEs simplify routing decisions and encourage expert specialization and robustness.

We define two structurally differing variants of MoE layers for CNNs to achieve a similar computational complexity regarding parameters but different computational structures: \textbf{BlockMoE} and \textbf{ConvMoE}. In \texttt{ResNet-18}, a BlockMoE replaces a BasicBlock, so each expert in a BlockMoE is a BasicBlock containing two convolutional layers and a skip connection. In \texttt{ResNet-50}, BlockMoE replaces a BottleneckBlock. A ConvMoE replaces a single convolutional layer in a BasicBlock or a BottleneckBlock. This way, each expert in a ConvMoE is a single convolutional layer. We further compare replacing convolutional layers and residual blocks individually and over the whole stage.

%It is a good choice for creating MoE layers that contain a large number of parameters in this network and can thus quickly increase the total amount of parameters without significant changes to the network architecture as \textit{layer4} itself already contains $8.4\cdot10^6$ out of the $11.7\cdot10^6$ full parameters of the CNN architecture.

\begin{figure}[t]
     \begin{subfigure}[t]{\columnwidth}
         \centering
         \includegraphics[width=\textwidth]{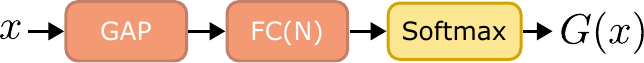}
         \caption{GAP-FC gate}
         \label{fig:gate_gapl}
     \end{subfigure}
     \begin{subfigure}[t]{\columnwidth}
         \centering
         \includegraphics[width=\textwidth]{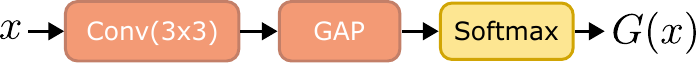}
         \caption{Conv-GAP gate}
         \label{fig:gate_cgap}
     \end{subfigure}
    \caption{Gate architectures: GAP-FC averages over each channel before applying any learned operation; the selection mechanism can not consider spatial details. The FC layer learns to select based on the averaged channels. Conv-GAP employs a convolution operation on the feature map, theoretically enabling it to learn spatially invariant patterns for each expert.}
    \label{fig:gates}
\end{figure}

\subsection{Gate}

The gating network, or gate, is responsible for activating and weighing the different experts.
%As gates are only responsible for mapping inputs to experts and are kept lightweight and fast.
%The experts do the actual computation.
%At the same time, this limits the ability to learn patterns in the input space and makes learning to select experts challenging.
We use two types of gating networks (see Figure~\ref{fig:gates}): the GAP-FC, introduced in~\cite{pavlitska2023sparsely}, and a Conv-GAP gate, which consists of a convolutional layer and a GAP layer. %The convolution layer outputs a channel for each expert. The following GAP layer averages over the expert's channel to compute its importance score. 
The convolution can use detailed, local information encoded in the input features but has to encode it so that the GAP of its output computes proper mappings to experts. 

%The computed importance scores can be used to activate any desired set of experts. In the sparse setting, the highest score is determined to activate the most essential expert. However, if multiple experts are activated, the scores are also used to weigh the experts. 

\subsection{Load Balancing Loss}

A balancing loss is essential in training MoE models to prevent routing collapse and encourage diverse expert utilization; common formulations include entropy-based losses that maximize expert assignment uncertainty and load balancing losses that promote uniform expert activation, such as those introduced by Shazeer et al.\cite{shazeer2017outrageously} and Riquelme et al.\cite{riquelme2021scaling}.

We transform the KL-loss $\loss{KL}$, introduced in~\cite{pavlitska2023sparsely} to allow another interpretation. With $\sum_{i=1}^{N}P(E_i)=1$ and the entropy of a vector $\mathcal{H}(v)=-\sum_{i=1}^{N}v_i\cdot\ln(v_i)$, we can derive
\begin{align*}
    \loss{KL}=D_{KL}(P ||Q) =\sum_{i=1}^{N}P(E_i)\cdot\ln\left(\frac{P(E_i)}{Q(E_i)}\right) \\
             =\sum_{i=1}^{N}P(E_i)\cdot\ln\left(\frac{P(E_i)}{1/N}\right) \\
             =-\mathcal{H}(P)+const =\loss{entropy}+const
             %&=\sum_{i=1}^{N}P(E_i)\cdot\left(\ln(P(E_i))+\ln(N)\right) \nonumber\\
             %&=\sum_{i=1}^{N}P(E_i)\cdot\left(\ln(P(E_i))+\ln(N)\right) \nonumber\\
             %&=\sum_{i=1}^{N}P(E_i)\cdot\ln(P(E_i))+\sum_{i=1}^{N}P(E_i)\ln(N) \nonumber\\
             % &=-\mathcal{H}(P)+const\nonumber \\
             % &=\loss{entropy}+const.
\end{align*}
We can thus use the entropy of the average expert importance vector $P$ as an equivalent entropy loss: $\loss{entropy}=-\mathcal{H}(P)$.

We also use the switch loss $\loss{switch}$~\cite{fedus2022switch}, which minimizes the cross-entropy between the empirical distribution of expert assignments and a uniform target distribution, encouraging balanced expert utilization while allowing sharper and less entropic routing decisions than entropy or KL-based losses.

Overall, entropy and KL losses promote balanced expert usage across the batch by maximizing the diversity of expert assignments, while switch loss operates at the input level to enforce high-entropy gating decisions, leading to softer and more exploratory routing per sample.

%We refer to this loss as $\loss{entropy}$ or less formally as the entropy loss.
%We can now view balancing loss in a theoretical information way.
% Minimizing $\loss{entropy}$ corresponds to maximizing the entropy of the importance vector of all experts, i.e., by reducing $\loss{KL}$, we minimize the prior knowledge to which expert an unknown input token will be routed. 

% We also use the \textit{switch loss} introduced in~\cite{fedus2022switch}. For an MoE layer with $N$ experts and a batch $\mathcal{B}$ of samples to be routed to the gate, the loss is defined as follows:

% \begin{equation*}
%     \loss{switch}(X) = \alpha \cdot N \cdot \sum_{i=1}^{N}(f_i \cdot P_i),
%     \label{eq:switch_loss}
% \end{equation*}

% Where $\alpha$ is a factor controlling the contribution to the overall loss, $f_i$ is the fraction of tokens dispatched to expert $i$:

% \begin{equation*}
%     f_i = \frac{1}{|\mathcal{B}|} \sum_{x \in \mathcal{B}} \mathbb{1} \left\{ \arg\max p(x) = i \right\},
% \end{equation*}

% and $P_i$ is the fraction of the router probability allocated for expert $i$ for all tokes in batch $\mathcal{B}$:

% \begin{equation*}
%     P_i = \frac{1}{|\mathcal{B}|} \sum_{x \in \mathcal{B}} p_i(x),
% \end{equation*}

% Where $p_i(x)$ is the probability of routing token $x$ to expert $i$.

\newpage
\section{Experiments and Evaluation}
We apply the proposed sparse MoE layers to CNNs for the image classification task and analyze the effect of the adversarial training. We further describe the experimental setting and the quantitative and qualitative results. 

\subsection{Experimental Setting}
\label{sec:classification_standard_training}

We use \texttt{ResNet-18} and \texttt{ResNet-50}~\cite{he2016deep}
models trained on \texttt{CIFAR-100}~\cite{krizhevsky2009learning}. We report the average precision on all classes calculated on the test split of the dataset.
%If not indicated otherwise, the evaluation is performed on the test split of the dataset. 
All experiments were performed using NVIDIA RTX4090.

Unless otherwise specified, we train each model for $200$ epochs using SGD with a momentum of $0.9$, a weight decay of $5\cdot10^{-4}$, and an initial learning rate of $10^{-2}$ following a polynomial decay schedule. % We use the standard training split of the \texttt{CIFAR-100} dataset and load training samples in batches of size $256$ before applying standard augmentation operations such as random cropping, flipping, and normalization.% \texttt{ResNet-18} baseline reached the accuracy of $73.0\%$.

We apply the $l_{\infty}$ limited PGD-20~\cite{madry2017towards} and AutoPGD~\cite{croce2020reliable} attacks and PGD-7 adversarial training. %The attack is started at a random position in a $l_{\infty}$ cube around the input with a side length of $8$ and runs for $20$ update steps using a step size of $2$.

%in the standard setting and using a limited PGD attack to adversarially perturb input images in the attacked setting to evaluate the model's adversarial robustness against any $l_{\infty}$ limited attacker. 
%Every experiment in the study modifies this setup on one or multiple axes.

\subsection{Performance, Adversarial Robustness, and Efficiency across Architectures}
\begin{figure}[t]
     \centering

     % \begin{subfigure}[b]{0.49\textwidth}
     %     \centering
         \includegraphics[width=\linewidth]{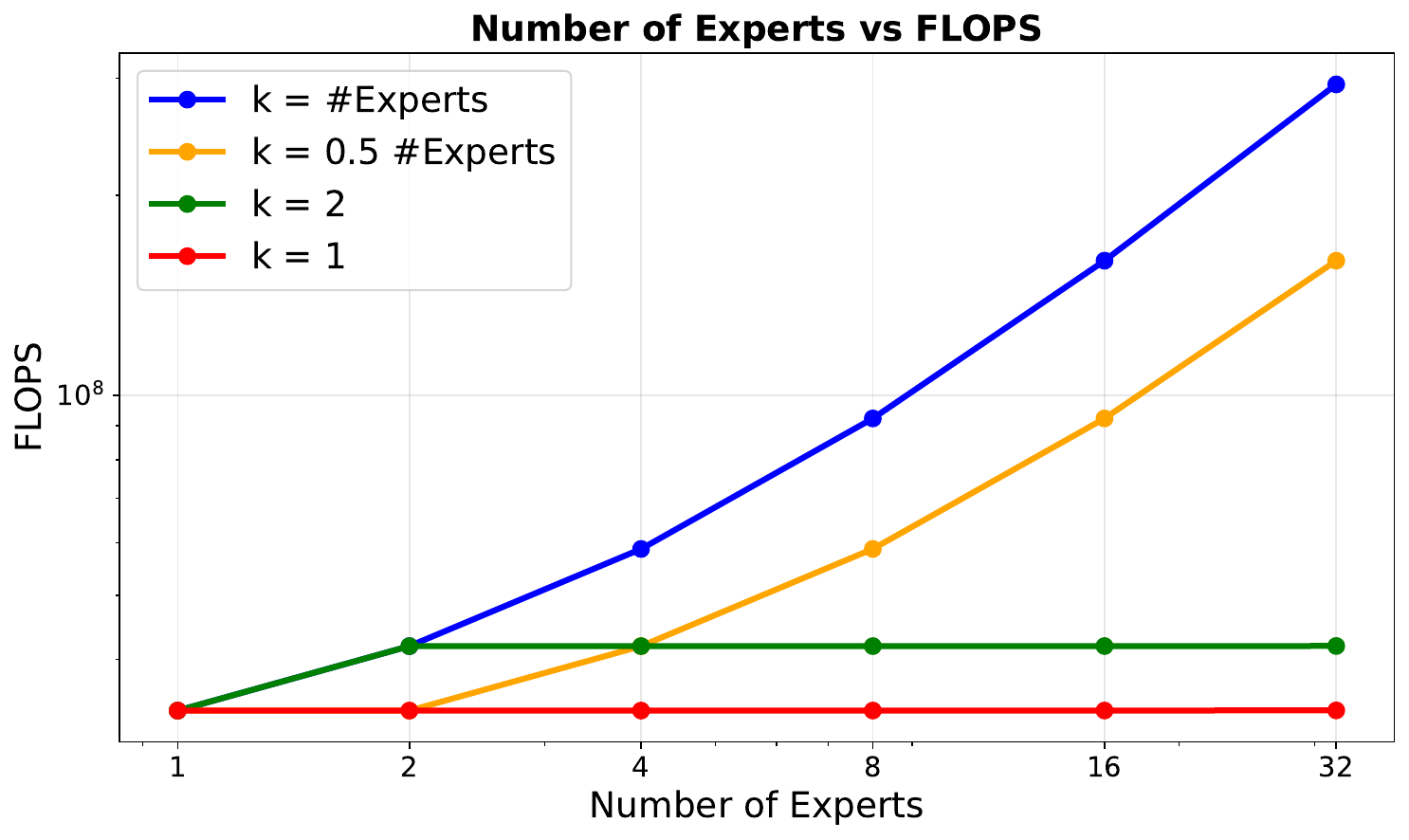}
     %     \caption{MoEBlock}
     %     \label{fig:flops_resnet_block_moe}
     % \end{subfigure}
     % \begin{subfigure}[b]{0.49\textwidth}
     %     \centering
     %     \includegraphics[width=\linewidth]{img/architectures/resnet_conv_moe_flops_params.png}
     %     \caption{ConvMoE}
     %     \label{fig:flops_resnet_conv_moe}
     % \end{subfigure}
        \caption{Computational cost of inference for the \texttt{ResNet-18} with a BlockMoE layer with $2-32$ experts per MoE layer.
        %Inference is conducted with images of size $28\times 28$ pixels and at different sparsity levels.
        The sparse case $k=1$ exhibits quasi-constant FLOPs when compared against models with half of all experts or even all experts activated. 
        }
        \label{fig:flops_resnet_moes}
\end{figure}

\textbf{Computational complexity:} The analysis of the impact of these architectures in terms of FLOPs and the number of parameters has shown, that if all experts are activated, the FLOPs grow almost at the same rate as the parameters (see Figure~\ref{fig:flops_resnet_moes}).
%The only reason it is not straight is the constant part of the network is unaffected by the scaling of the number of experts.
Qualitatively, the same behavior is shown when half of all experts are active.
However, if the number of activated experts $k$ is decoupled from the total number of experts in the MoE layers, the parameter counts decouple from the FLOPs.
This scaling behavior motivates the usage of sparse MoE layers in this work. %Knowing this scaling behavior is essential, and understanding that it essentially allows scaling models to enormous sizes without requiring more computing.

\begin{figure*}[t]
\centering

      \centering
      \includegraphics[width=0.5\linewidth]{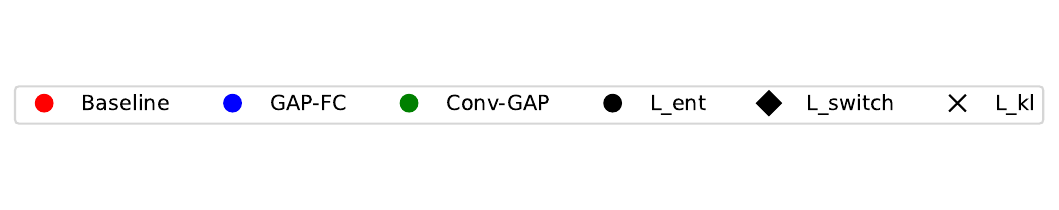}

  \begin{subfigure}{\linewidth}
      \centering
      \includegraphics[width=0.246\linewidth]{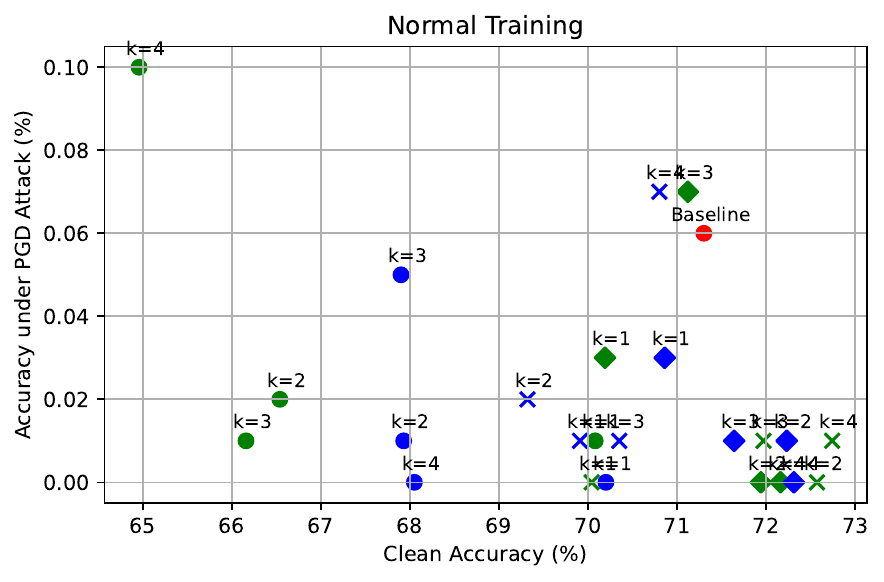}
      \includegraphics[width=0.246\linewidth]{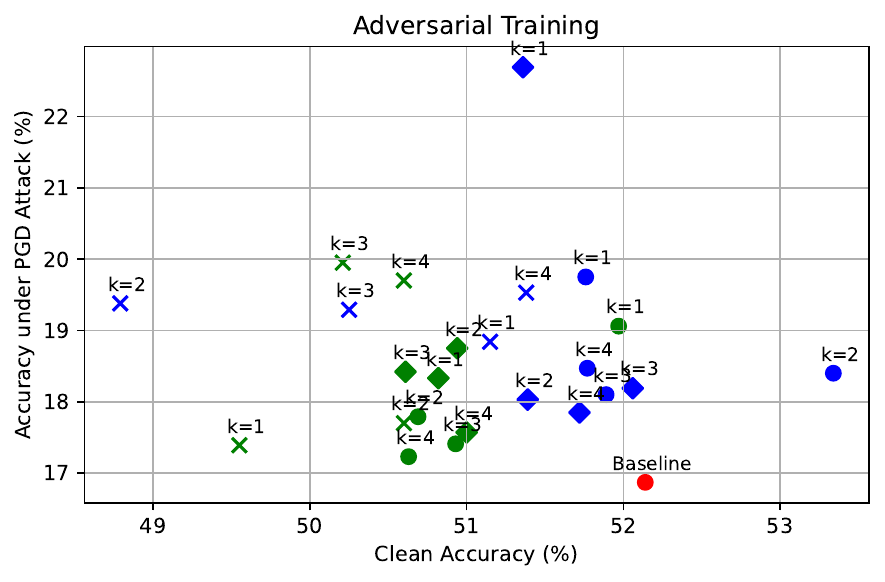}
      \includegraphics[width=0.246\linewidth]{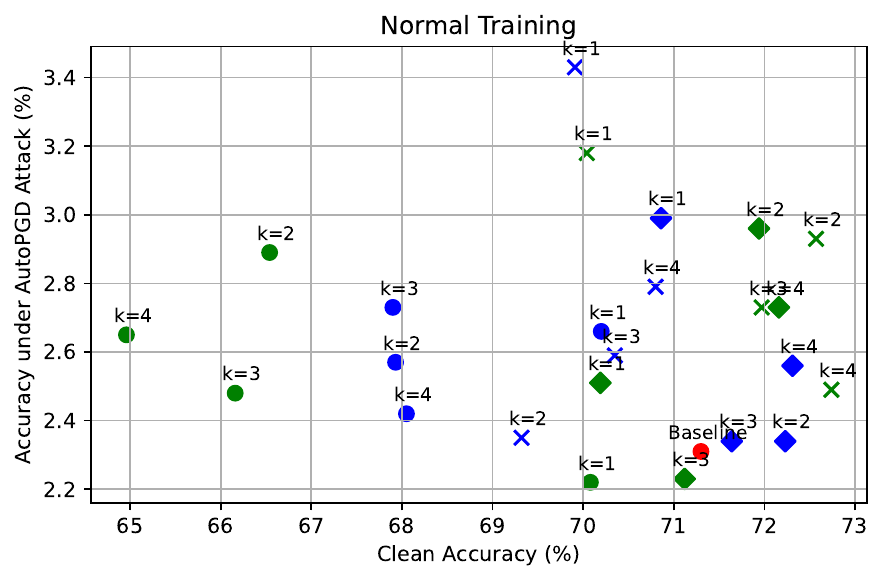}
      \includegraphics[width=0.246\linewidth]{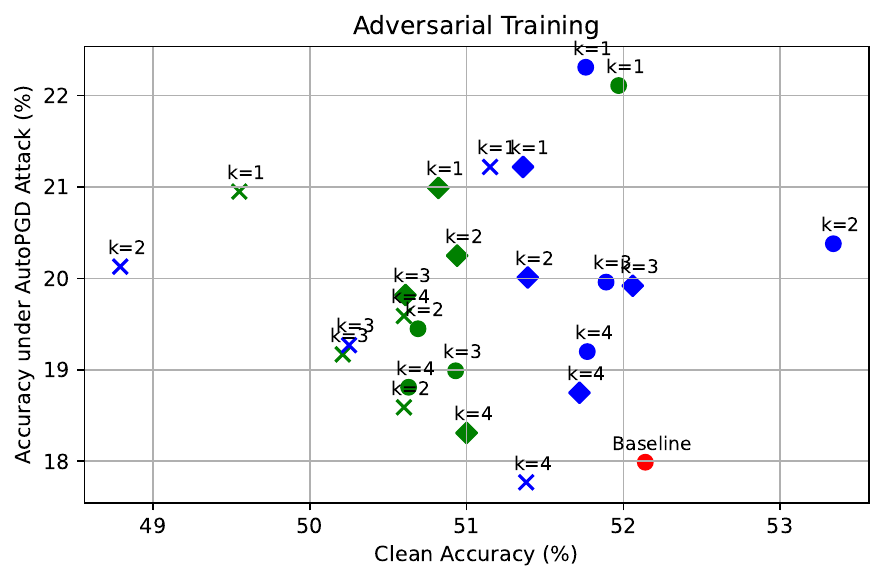}
      \caption*{\texttt{ResNet-18}, BlockMoE}
    \end{subfigure}
    
    \begin{subfigure}{\linewidth}
      \centering
      \includegraphics[width=0.246\linewidth]{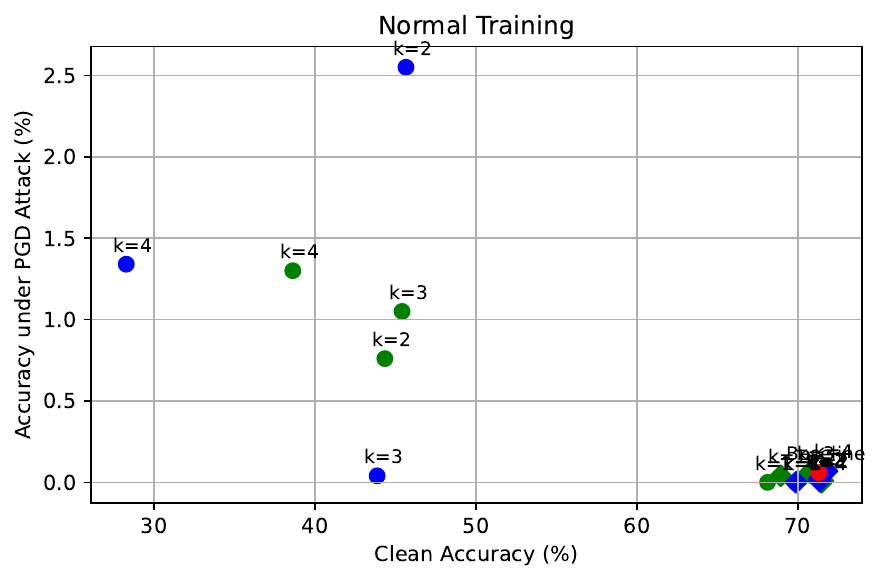}
      \includegraphics[width=0.246\linewidth]{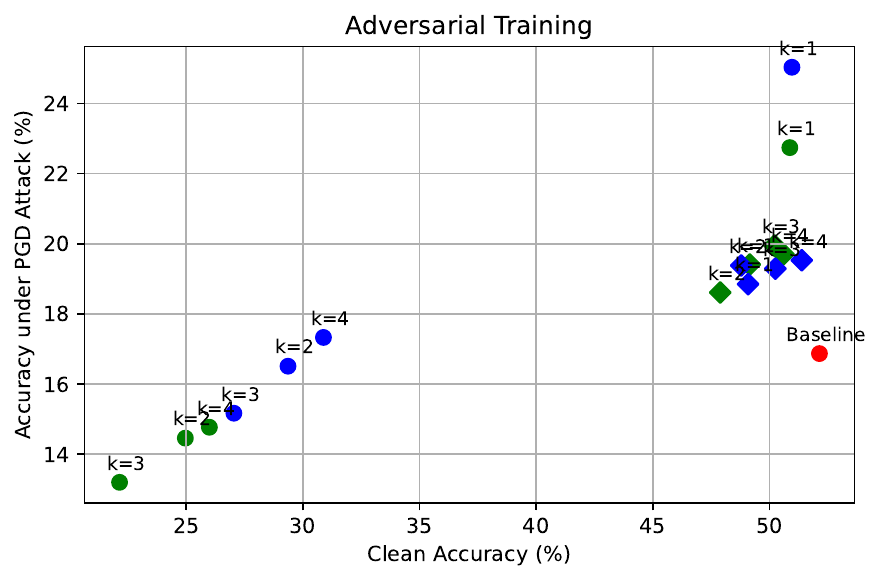}
      \includegraphics[width=0.246\linewidth]{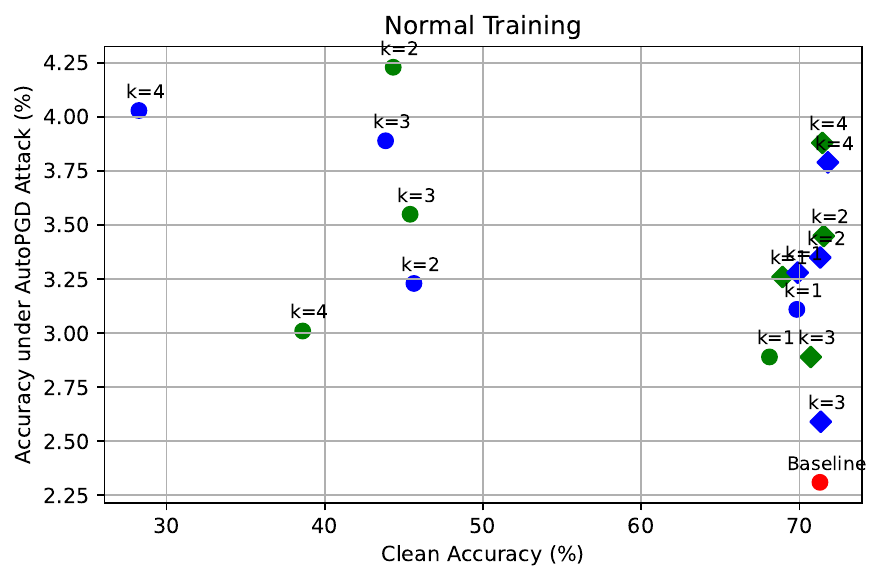}
      \includegraphics[width=0.246\linewidth]{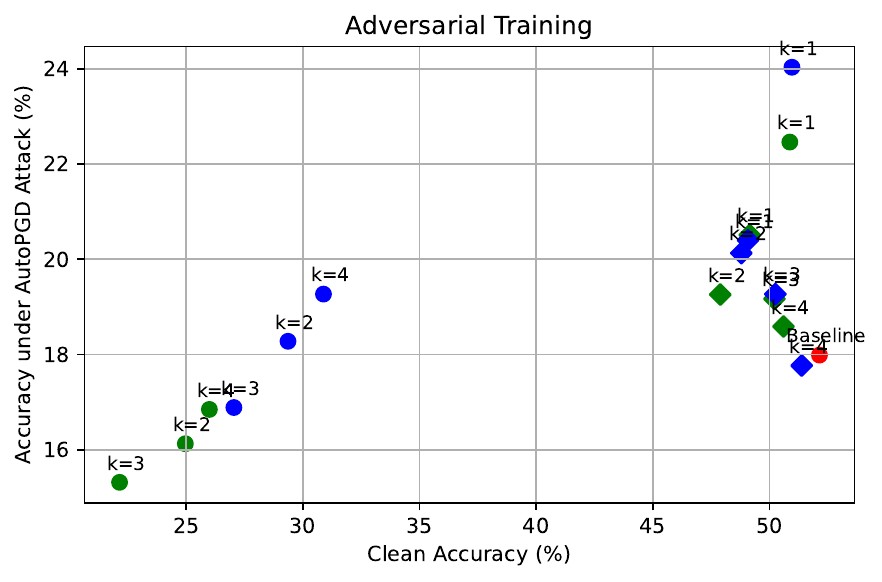}
      \caption*{\texttt{ResNet-18}, ConvMoE}
    \end{subfigure}

    \begin{subfigure}{\linewidth}
      \centering
      \includegraphics[width=0.246\linewidth]{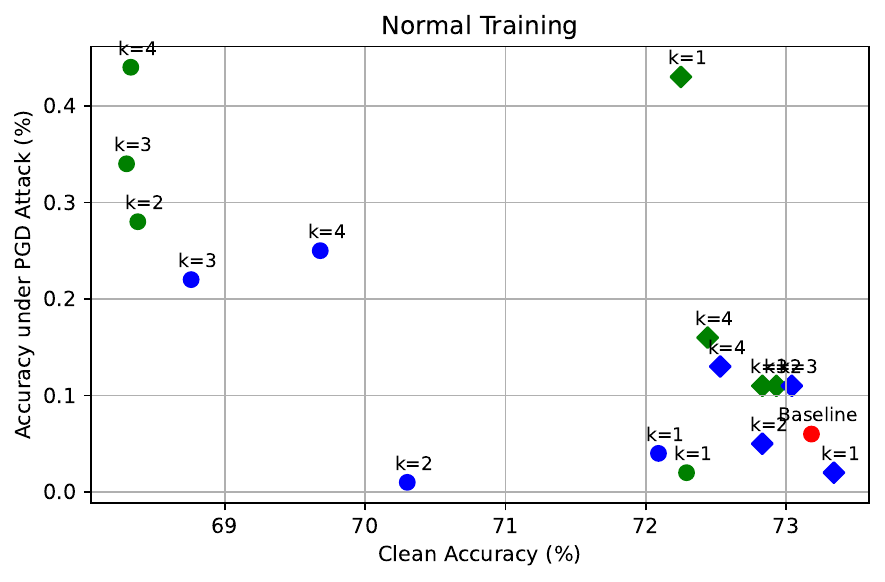}
      \includegraphics[width=0.246\linewidth]{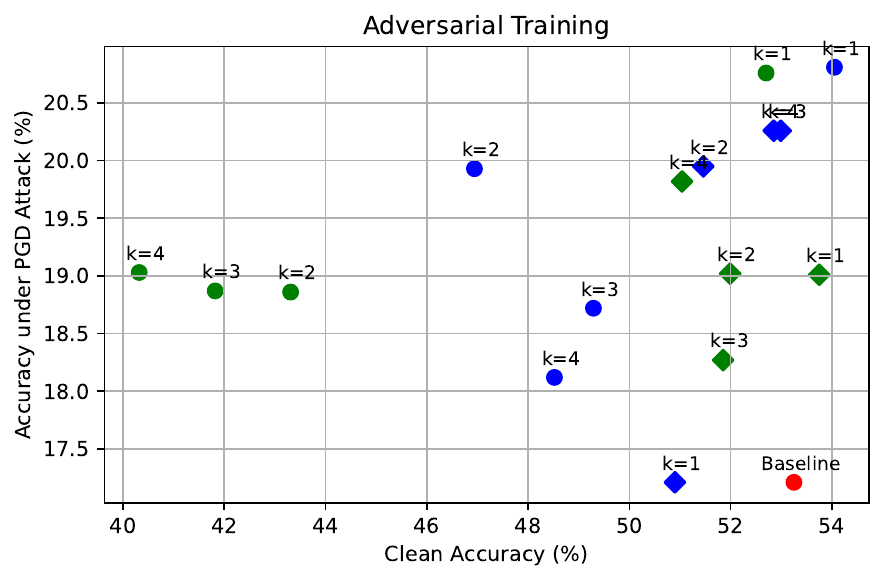}
      \includegraphics[width=0.246\linewidth]{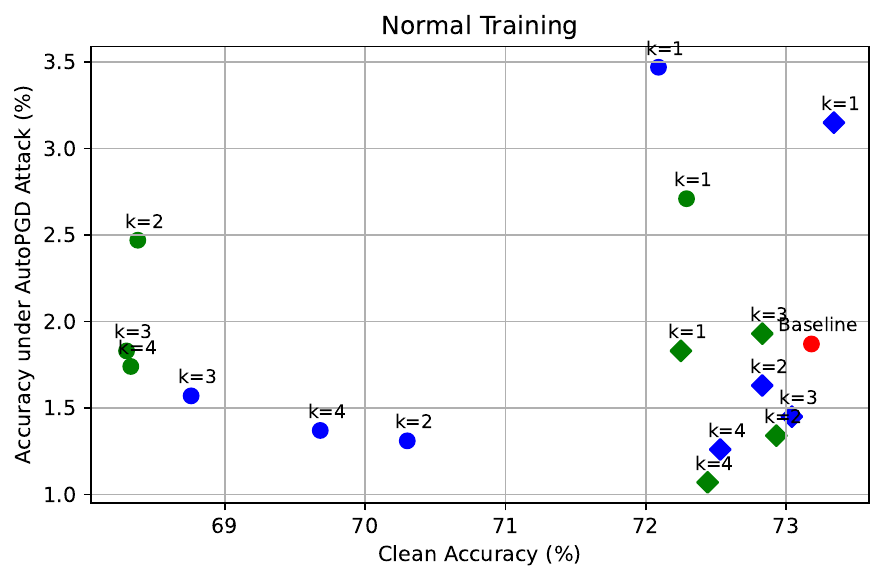}
      \includegraphics[width=0.246\linewidth]{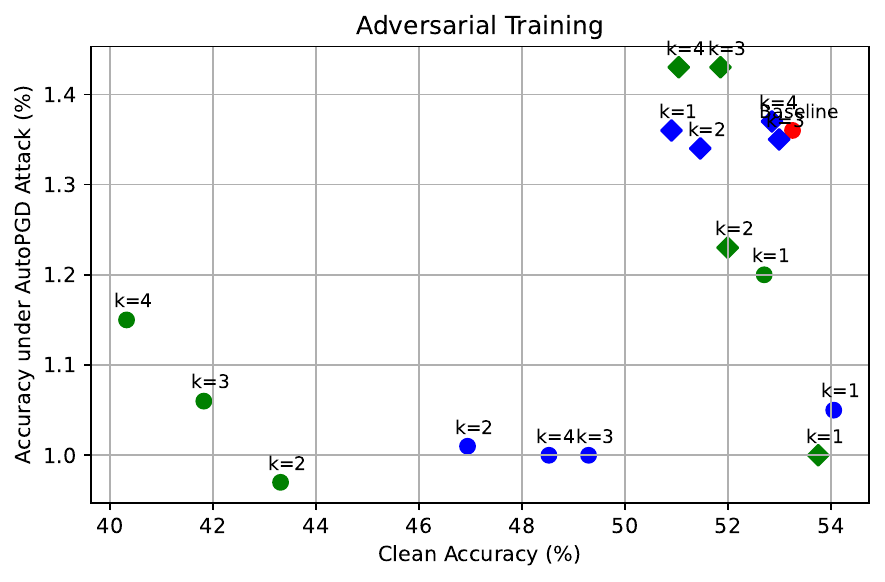}
      \caption*{\texttt{ResNet-50}, BlockMoE}
    \end{subfigure}

    \begin{subfigure}{\linewidth}
      \centering
      \includegraphics[width=0.246\linewidth]{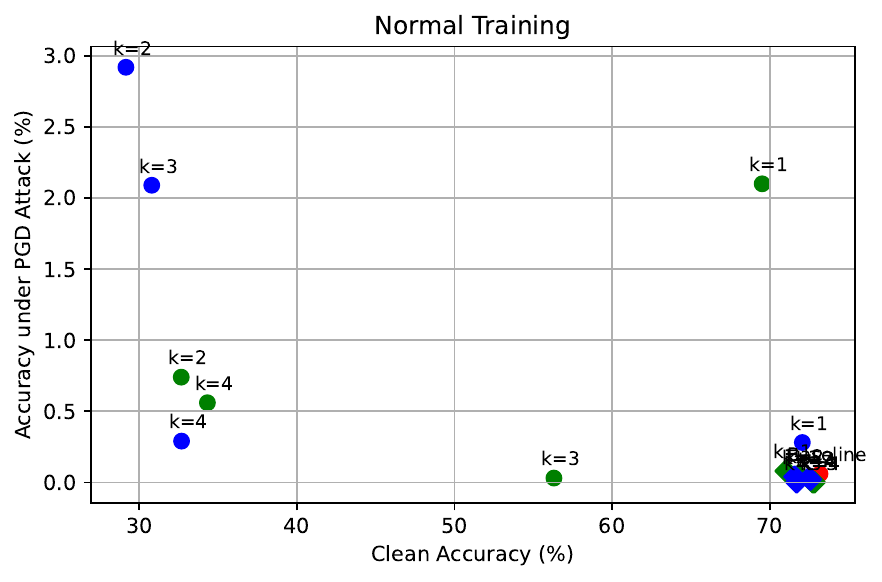}
      \includegraphics[width=0.246\linewidth]{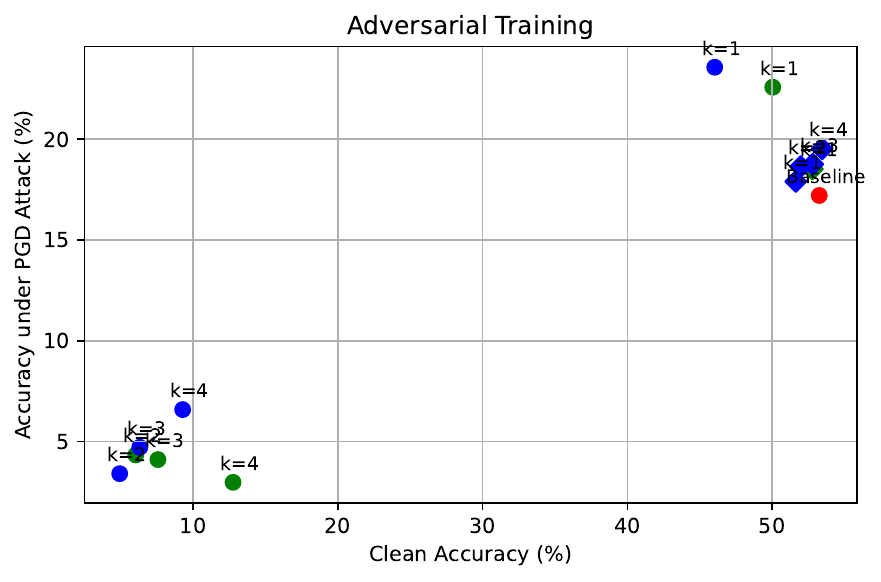}
      \includegraphics[width=0.246\linewidth]{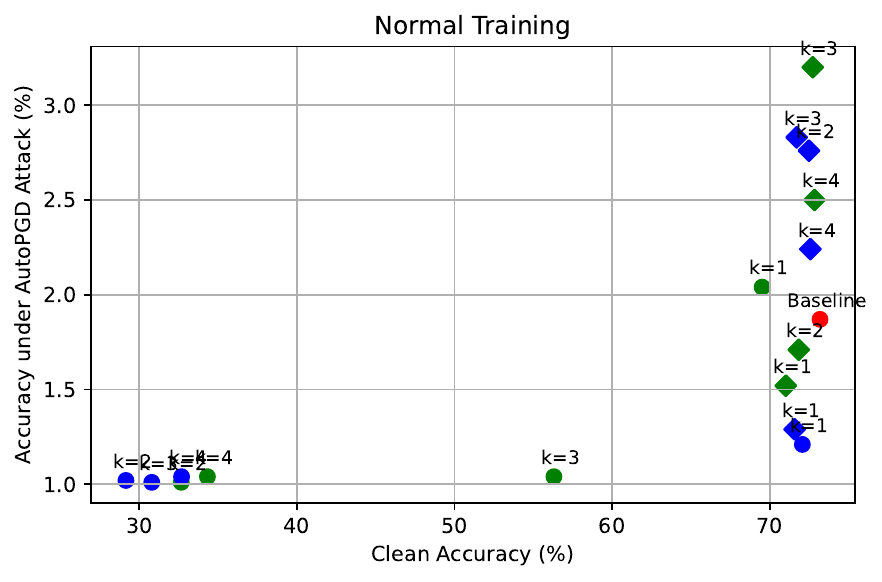}
      \includegraphics[width=0.246\linewidth]{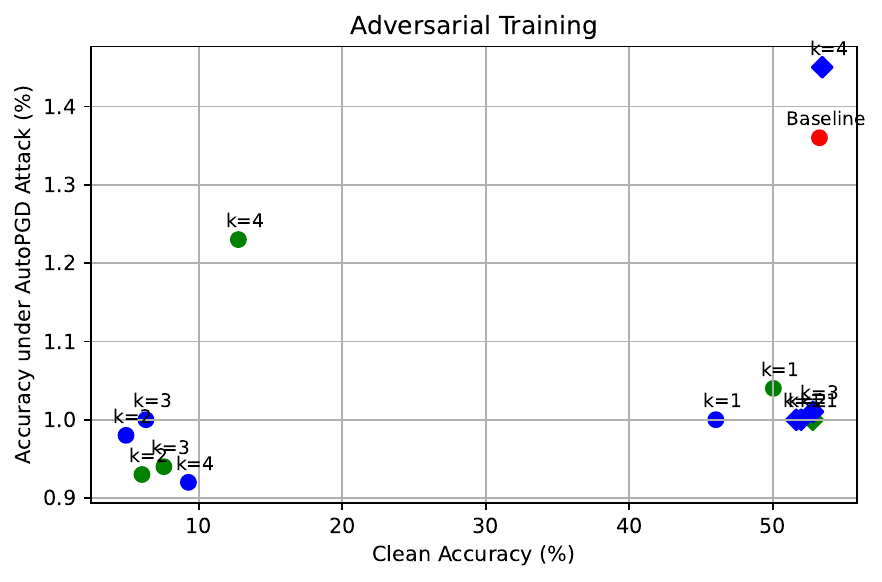}
      \caption*{\texttt{ResNet-50}, ConvMoE}
    \end{subfigure}

\caption{Robustness-accuracy trade-off for the normally and adversarially trained baselines and models with MoE layers 2 BlockMoE layers, each replacing a BasicBlock in the \textit{conv5\_x}. MoE layers have 4 experts and a varying $k$ for the top-k routing. Each subplot shows clean accuracy (x-axis) vs. accuracy under PGD or AutoPGD attack (y-axis). Architectures that outperform the baseline lie to the top or right of the red dot in each plot. Higher placement of the dots for MoE models relative to the baseline indicates improved adversarial robustness, and further to the right indicates better clean accuracy.}
\label{fig:tradeoff}
\end{figure*}

\textbf{Comparison to the baseline:} Across both \texttt{ResNet-18} and \texttt{ResNet-50} architectures, adversarial training consistently leads to larger improvements in robustness when MoE layers are introduced (see Figure \ref{fig:tradeoff}). Under normal training, both ConvMoE and BlockMoE models show only modest gains in adversarial accuracy, with little to no improvement in clean accuracy compared to the baseline. In contrast, adversarially trained MoE models outperform the baseline significantly in robustness, while also maintaining or even slightly improving clean accuracy. This effect is more pronounced in \texttt{ResNet-50}, suggesting that deeper models benefit more from input-dependent expert routing under adversarial attacks.

\textbf{ConvMoE vs. BlockMoE:} Models with BlockMoE layers demonstrate more consistent improvement in both robustness and clean accuracy, especially under adversarial training. ConvMoE models show competitive performance in some cases, particularly with \texttt{ResNet-18}, but their improvements are less stable across training regimes and datasets. BlockMoEs provide a higher-level structural adaptation that appears better suited for expert specialization, leading to stronger robustness–accuracy tradeoffs in most settings. We further analyze the impact of the selected loss on individual experts in Section~\ref{subsec:analysis_individual}.

\textbf{Best-performing setup:} For both \texttt{ResNet-18} and \texttt{ResNet-50}, the best-performing MoE configurations consistently use Top-2 routing, entropy-based auxiliary loss, and GAP-FC gating. These models lie clearly above the dense baseline in the tradeoff plots, achieving higher robustness under PGD and AutoPGD attacks while preserving or even improving clean accuracy. This combination enables the gating network to make semantically meaningful decisions while encouraging expert specialization, which proves most effective under adversarial training conditions.

\textbf{Impact of $k$}: The number of activated experts per input $k$ plays a key role in balancing accuracy and robustness. Across both architectures and MoE types, increasing k from 1 to 2 consistently improves robust accuracy, especially under adversarial training. However, the marginal gains taper off beyond $k=2$, and in some cases, higher $k$ slightly reduces clean accuracy due to reduced sparsity. Overall, Top-2 routing offers the best tradeoff, enabling sufficient expert diversity without sacrificing sparsity-driven benefits like regularization and efficiency.

% Figure~\ref{fig:classification_num_experts} shows the standard and adversarial accuracy against a PGD attacker for various experts and models trained with entropy and switch loss.
% The first plot shows a slight anti-correlation between the number of experts and accuracy for both losses. There is low or even vanishing accuracy under the PGD attack. The MoE models are thus not a viable option for increasing accuracy.
% Under the assumption that the additional parameters are not beneficial for the accuracy of these models, adding more experts to the models is expected to decrease the overall performance as activations are more sparse. Each expert sees fewer tokens during training.

%Having seen robustness improvements in all of the MoE variants, from sparse to fully activated, we now want to revert our attention to the completely sparse MoEs.
%We again train models with a different number of experts but train them using PGD-7 AT.  %shows both the standard and adversarial accuracy against a PGD-20 attacker.

\textbf{Balancing loss:} Across the robustness–accuracy trade-off graphs, entropy loss consistently outperforms the switch loss in terms of adversarial robustness. Models trained with the entropy loss achieve higher PGD and AutoPGD accuracy in nearly all configurations, particularly under adversarial training. Models trained with the entropy loss (shown as circles in Figure \ref{fig:tradeoff}) dominate the upper-right regions of the graphs, especially when paired with Top-2 routing. These results suggest that the entropy loss encourages more effective expert specialization and routing, leading to improved robustness without sacrificing clean accuracy. The KL loss outperforms the entropy loss under normal training, but shows worse results under adversarial training.

\begin{figure}[t]
     \centering
     \includegraphics[width=\linewidth]{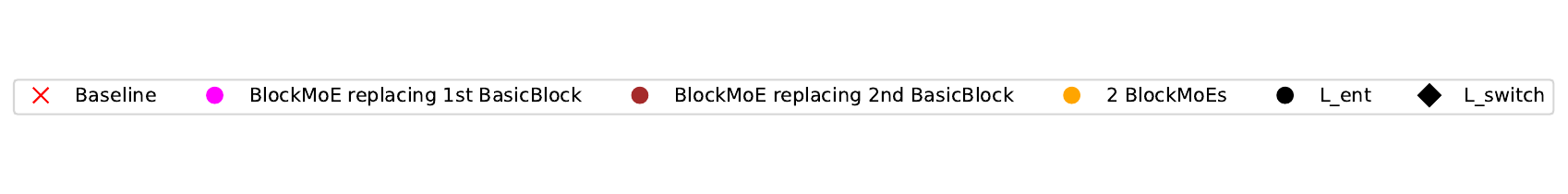}
     
     \begin{subfigure}{0.49\linewidth}
      \centering
      \includegraphics[width=\linewidth]{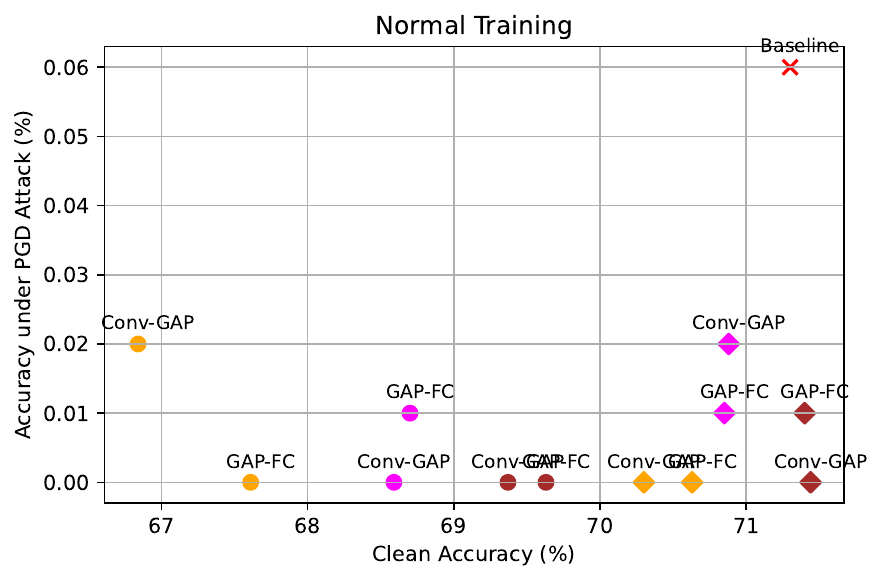}
      \caption*{$conv4\_x$, PGD, normal training}
    \end{subfigure}
    \hfill
    \begin{subfigure}{0.49\linewidth}
      \centering
      \includegraphics[width=\linewidth]{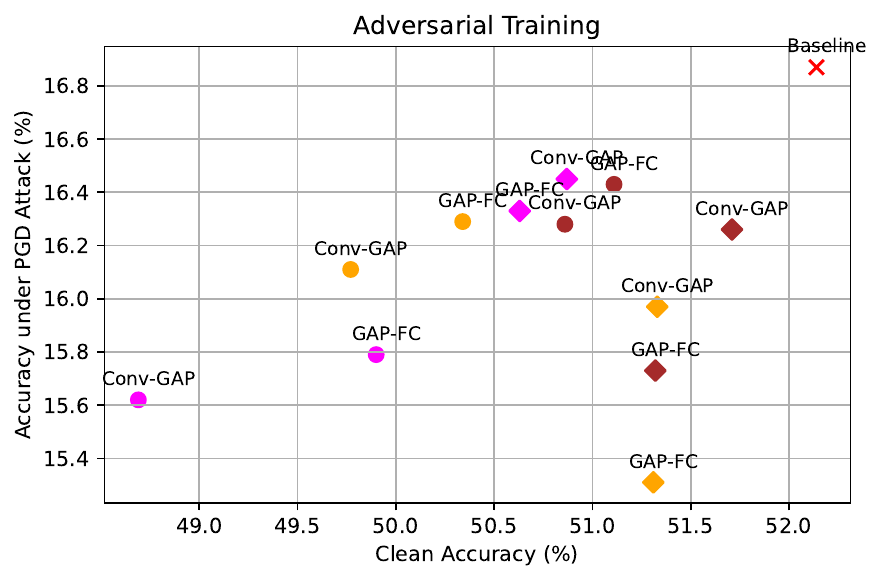}
      \caption*{$conv4\_x$, PGD, adversarial training}
    \end{subfigure}
    \hfill

    \begin{subfigure}{0.49\linewidth}
      \centering
      \includegraphics[width=\linewidth]{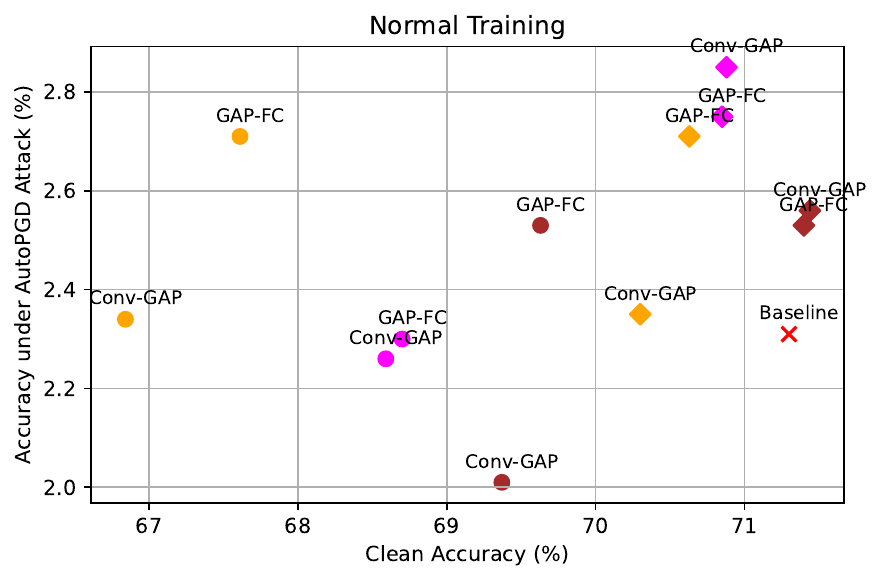}
      \caption*{$conv4\_x$, AutoPGD, normal training}
    \end{subfigure}
    \hfill
    \begin{subfigure}{0.49\linewidth}
      \centering
      \includegraphics[width=\linewidth]{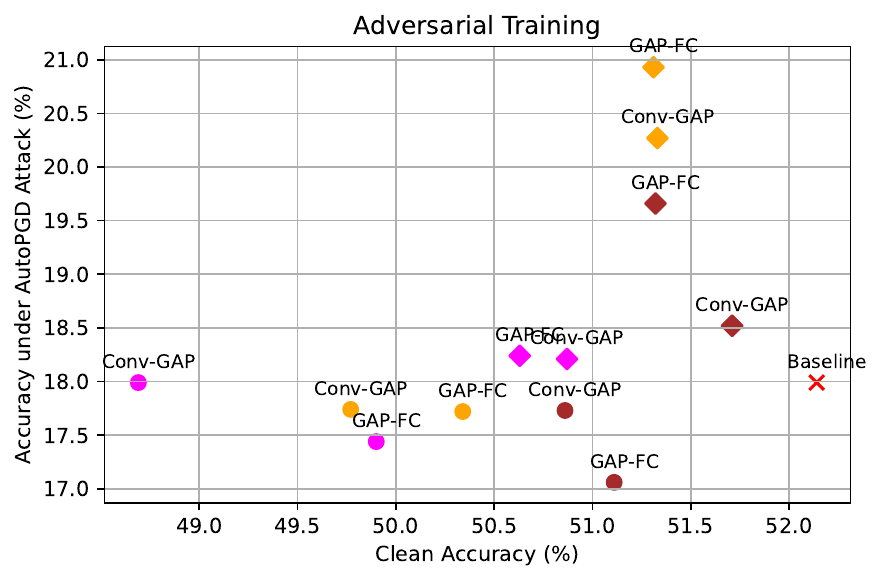}
      \caption*{$conv4\_x$, AutoPGD, adversarial training}
    \end{subfigure}
    \hfill

    \begin{subfigure}{0.49\linewidth}
      \centering
      \includegraphics[width=\linewidth]{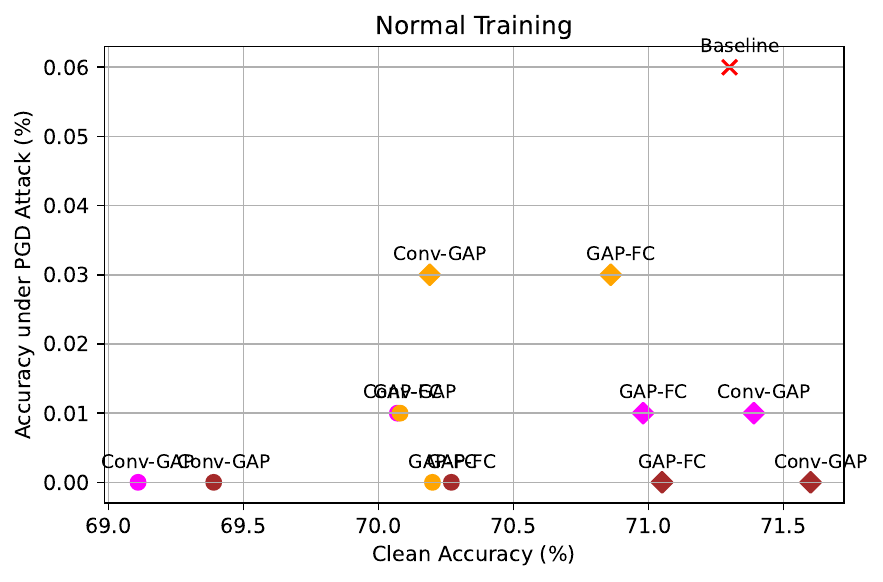}
      \caption*{$conv5\_x$, PGD, normal training}
    \end{subfigure}
    \hfill
    \begin{subfigure}{0.49\linewidth}
      \centering
      \includegraphics[width=\linewidth]{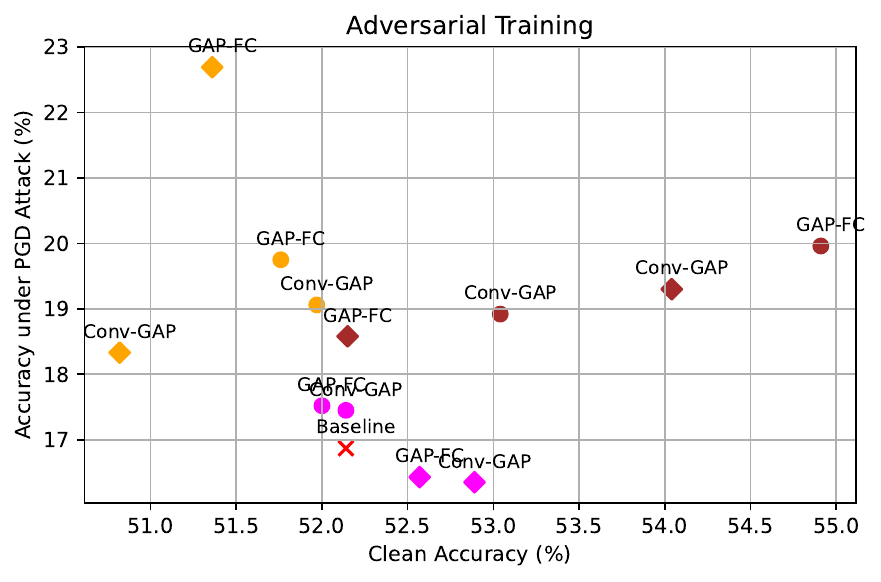}
      \caption*{$conv5\_x$, PGD, adversarial training}
    \end{subfigure}
    \hfill

    \begin{subfigure}{0.49\linewidth}
      \centering
      \includegraphics[width=\linewidth]{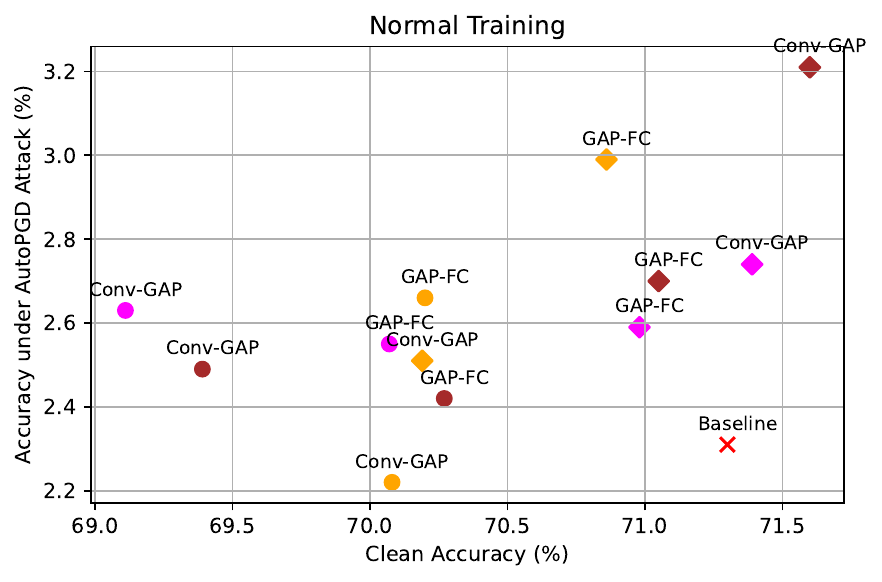}
      \caption*{$conv5\_x$, AutoPGD, normal training}
    \end{subfigure}
    \hfill
    \begin{subfigure}{0.49\linewidth}
      \centering
      \includegraphics[width=\linewidth]{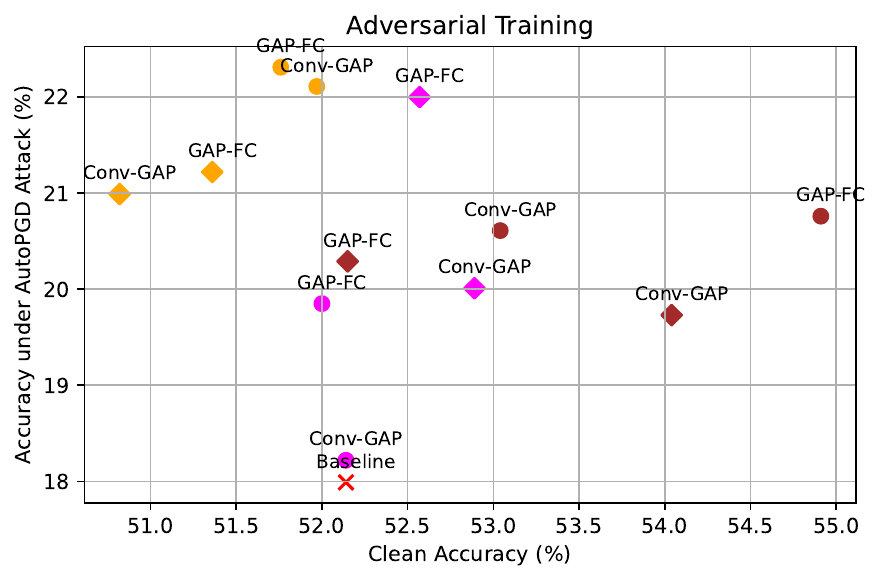}
      \caption*{$conv5\_x$, AutoPGD, adversarial training}
    \end{subfigure}
    \hfill
    
    \caption{Robustness-accuracy trade-off for different positions and number of BlockMoE layers in \texttt{ResNet-18}. MoE layers have 4 experts and use top-1 routing ($k=1$). Other settings follow Figure~\ref{fig:tradeoff}.}
        \label{fig:ablation-position-block}
\end{figure}

\textbf{Position of MoE layers:} Prior works~\cite{rajbhandari2022deepspeed,riquelme2021scaling,fedus2022switch} suggest placing MoE layers deeper in the network to achieve task-specific specialization and robustness. Therefore, we evaluate placing the proposed MoE layers in the deep $conv\_4x$ and $conv\_5x$ stages. The ablation studies for the BlockMoE (see Figure~\ref{fig:ablation-position-block}) reveal that the most effective architectural choice for improving robustness and maintaining clean accuracy is to replace the second BasicBlock in $conv5\_x$ with a BlockMoE layer. This setup (brown dots in Figure~\ref{fig:ablation-position-block}) consistently outperforms the baseline under adversarial training for PGD and AutoPGD attacks, achieving higher clean and adversarial accuracy. In contrast, using two BlockMoEs (orange) does not consistently yield better results and can introduce a drop in clean performance, suggesting diminishing returns from overusing experts. Also, the placement and number of BlockMoE layers have a larger impact on robustness and accuracy than the choice of loss function or gating mechanism. 

\begin{figure}[t]
     \centering
     \includegraphics[width=\linewidth]{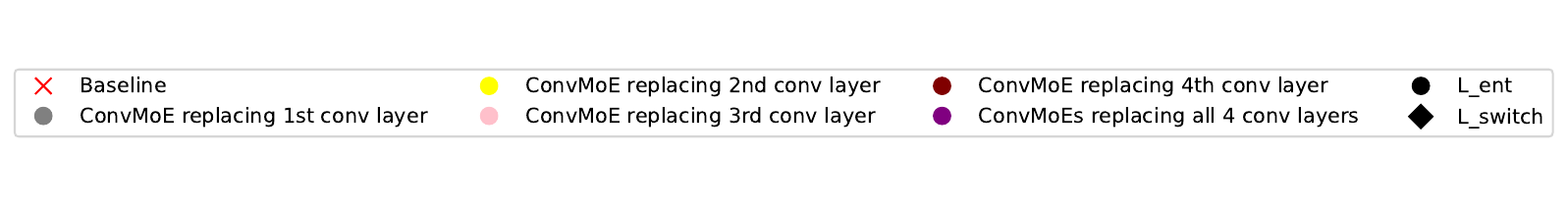}

    \begin{subfigure}{0.49\linewidth}
      \centering
      \includegraphics[width=\linewidth]{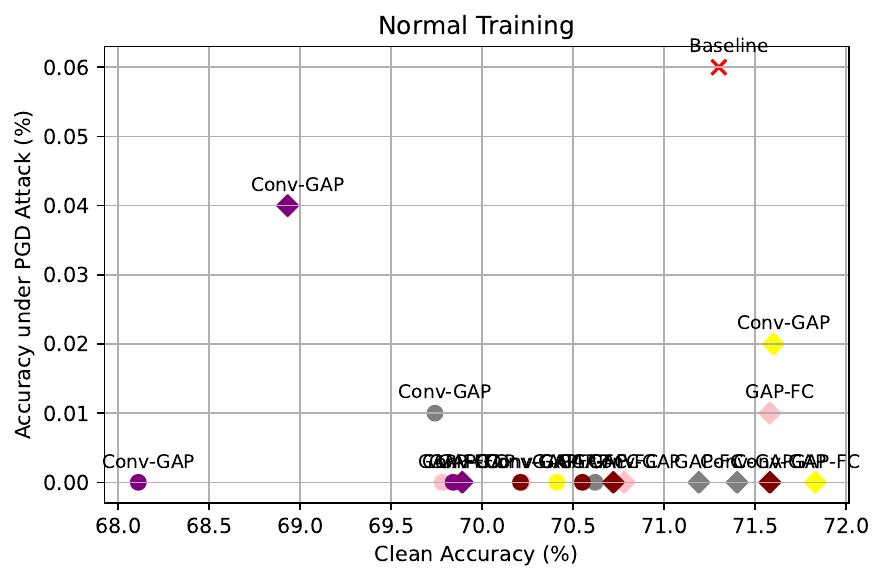}
      \caption*{$conv5\_x$, PGD, normal training}
    \end{subfigure}
    \hfill
    \begin{subfigure}{0.49\linewidth}
      \centering
      \includegraphics[width=\linewidth]{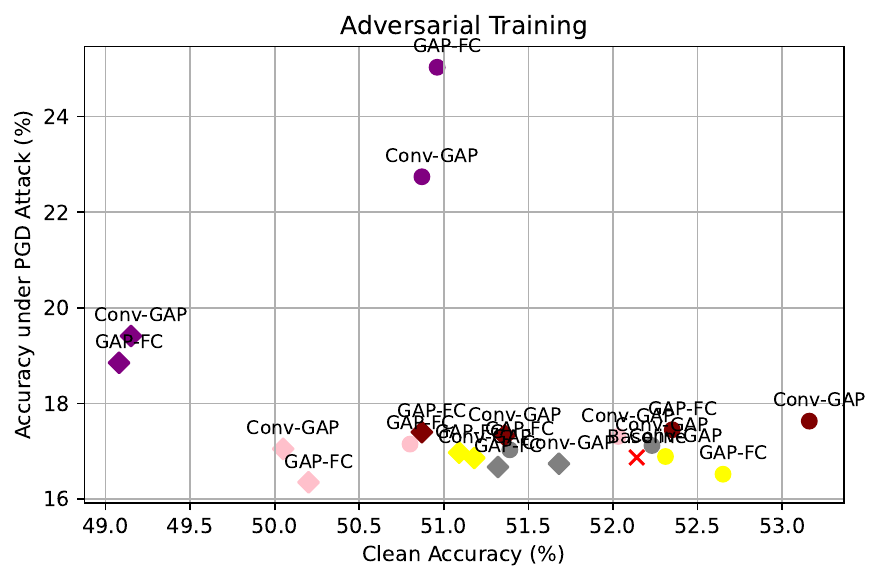}
      \caption*{$conv5\_x$, PGD, adversarial training}
    \end{subfigure}
    \hfill

    \begin{subfigure}{0.49\linewidth}
      \centering
      \includegraphics[width=\linewidth]{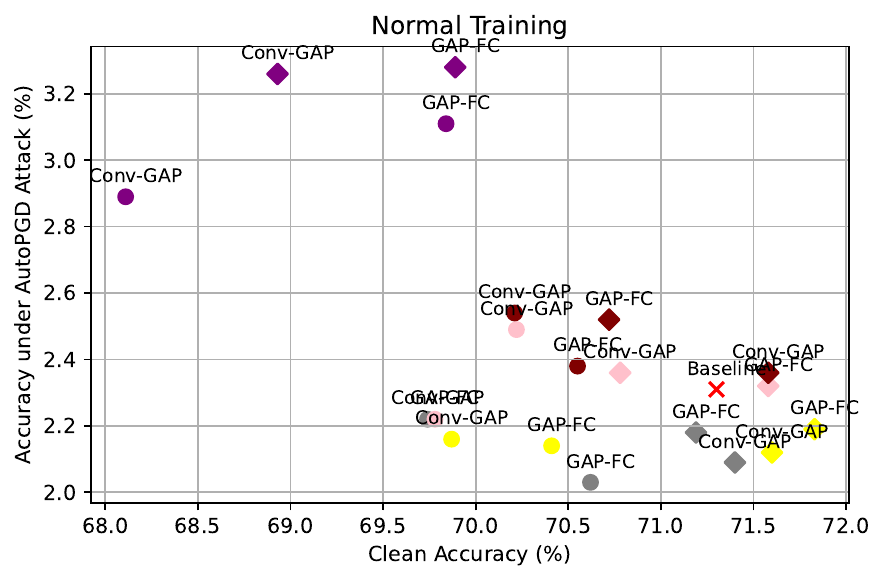}
      \caption*{$conv5\_x$, AutoPGD, normal training}
    \end{subfigure}
    \hfill
    \begin{subfigure}{0.49\linewidth}
      \centering
      \includegraphics[width=\linewidth]{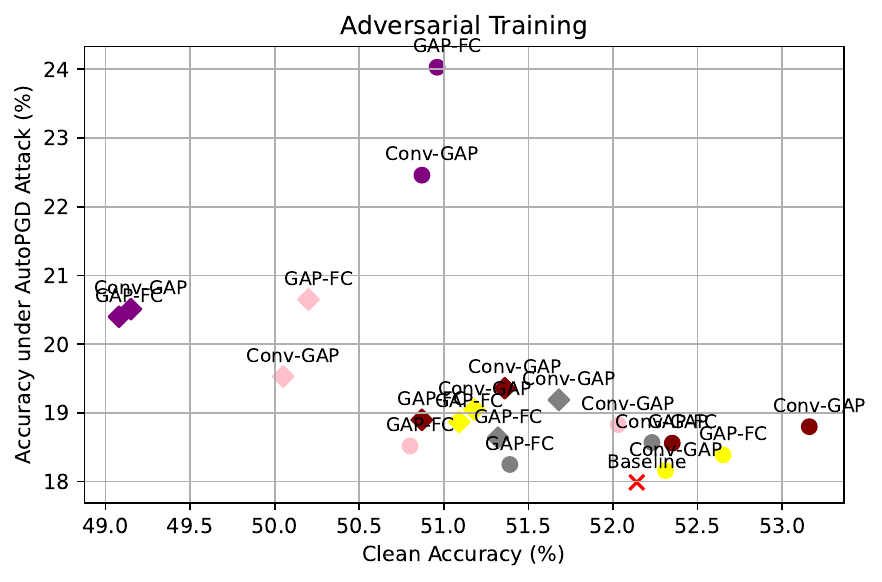}
      \caption*{$conv5\_x$, AutoPGD, adversarial training}
    \end{subfigure}
    \hfill
    
    \caption{Robustness-accuracy trade-off for different positions and number of ConvMoE layers in \texttt{ResNet-18}. MoE layers have 4 experts and use top-1 routing ($k=1$). Other settings follow Figure~\ref{fig:tradeoff}.}
        \label{fig:ablation-position-conv}
\end{figure}

Similarly, for ConvMoE (see Figure~\ref{fig:ablation-position-conv}), placing a single ConvMoE layer at the fourth convolutional position in the $conv5\_x$ stage leads to the best results, compared to replacing other layers or all layers.  Similar to the BlockMoE case, the impact of loss and gating choice is relatively minor.

\textbf{Number of Experts:}  To study the effect of expert count, we fix the architecture to \texttt{ResNet-18} with a BlockMoE layer and a Conv-GAP gate, and vary the number of experts per MoE layer: 1, 2, 4, 8, 16, and 32 (see blue solid line in Figure~\ref{fig:fixed-expert-plots}). The largest model has approximately $2.7\cdot 10^7$ parameters, about 23 times more than the base \texttt{ResNet-18}.

Under PGD-7 adversarial training, standard accuracy drops when using more than eight experts. Models trained with switch loss slightly improve up to eight experts, while those trained with entropy loss show a consistent decline as the expert count increases.

In terms of adversarial robustness, switch and entropy losses lead to improved performance as the number of experts increases up to eight, after which robustness declines. The decline in adversarial accuracy at high expert counts might indicate that the training quality degrades for the individual experts due to the high sparsity and the low number of tokens each expert receives. 

Overall, increasing the number of experts tends to improve both clean and adversarial accuracy up to a moderate point (typically 8–16 experts), beyond which performance levels off or decreases. The pattern is more noticeable with entropy loss and adversarial training, where expert diversity appears to have a stronger effect on robustness.

\subsection{Analysis of Individual Experts}
\label{subsec:analysis_individual}
We analyze the behavior of individual experts, focusing on routing and the effect of fixing the expert during inference, to investigate routing collapse, utilization patterns, and their implications for robustness and specialization.

%We focus on a single type of architecture and vary the number, the training procedure, and t, as well as the auxiliary loss used during training.
%To be precise, we investigate some of the previously trained BlockMoE architecture and train it for 200 epochs using the PGD-7 AT method.
%We investigate variants with
%\begin{equation}
%    N=\left[2,4,8,16,32\right]
%\end{equation}
%experts.
%These models range in their parameter counts from about $11$ million all the up to $270$ million and thus allow us to study models of structurally the same architecture but vary by more than one order of magnitude in their parameter counts.
%We start by investigating models trained on the CIFAR100 dataset using the BlockMoE and ConvMoE architectures. 

\begin{figure*}
\centering
\begin{subfigure}{\linewidth}
    \begin{subfigure}{.246\linewidth}
      \centering
      \includegraphics[width=\linewidth]{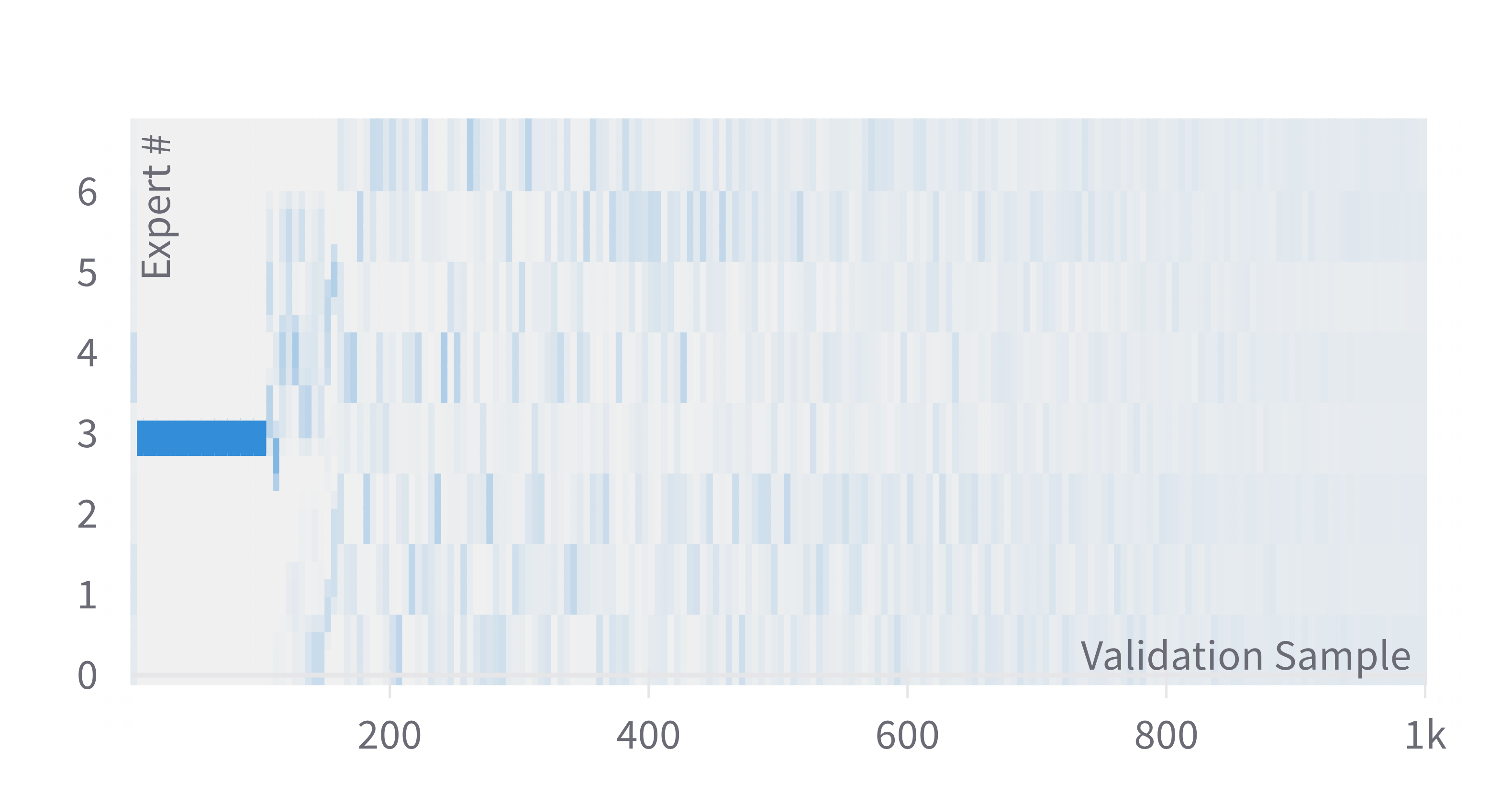}
      \caption{Normal training, switch loss}
    \end{subfigure}
    \hfill
    \begin{subfigure}{.246\linewidth}
      \centering
      \includegraphics[width=\linewidth]{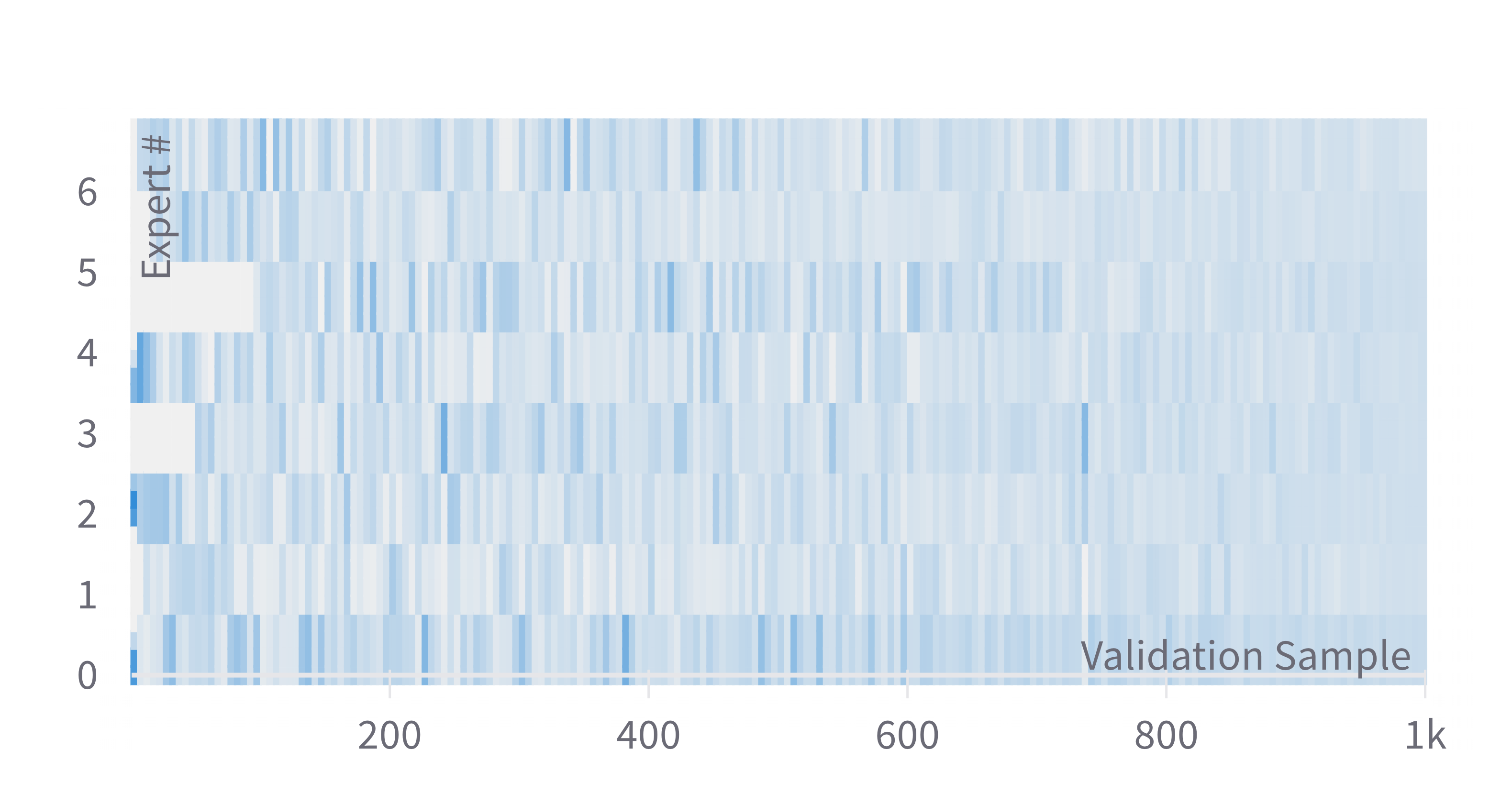}
      \caption{Normal training, entropy loss}
    \end{subfigure}
    \hfill
    \begin{subfigure}{.246\linewidth}
      \centering
      \includegraphics[width=\linewidth]{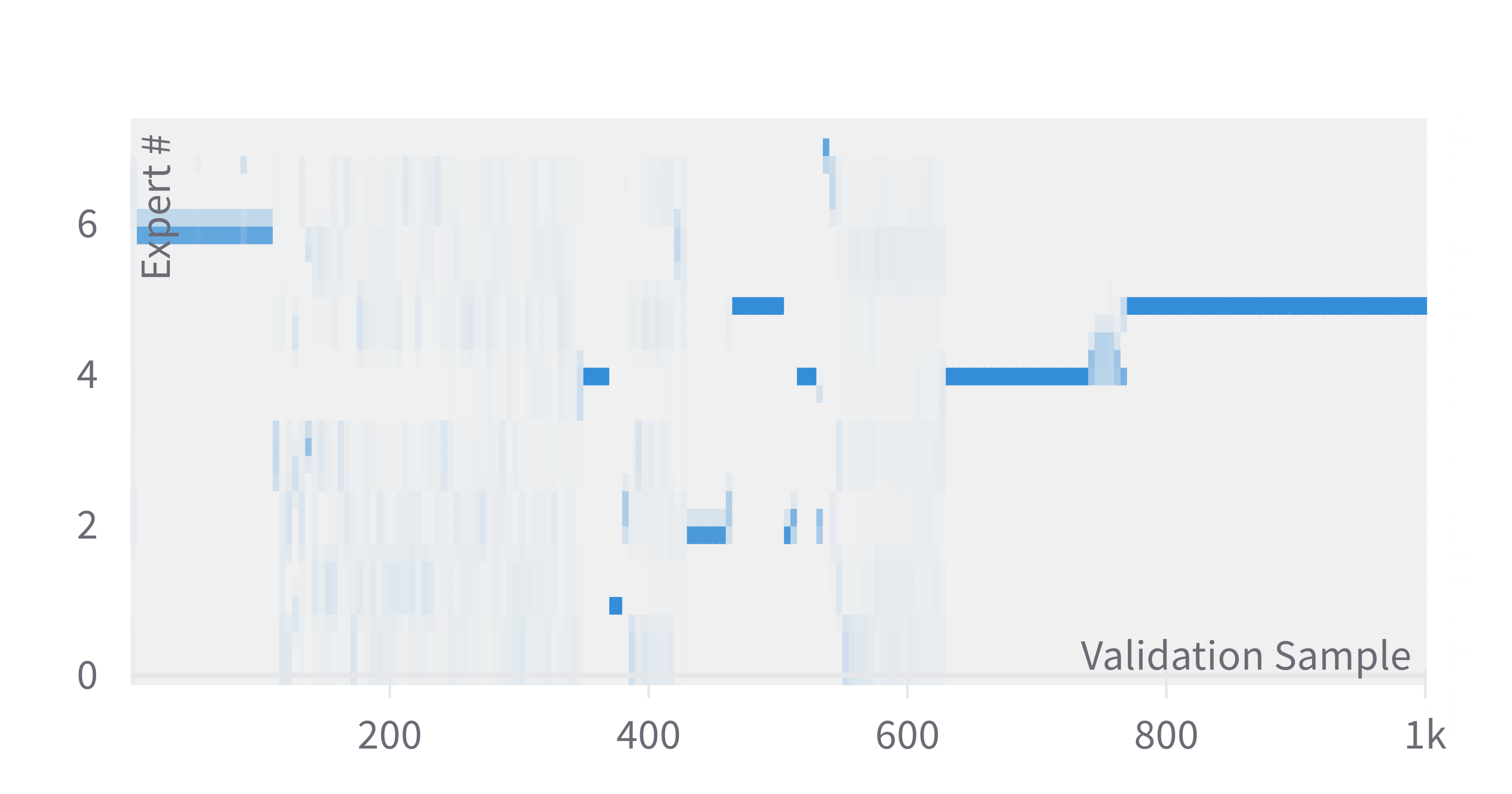}
      \caption{Adversarial training, switch loss}
    \end{subfigure}
    \hfill
    \begin{subfigure}{.246\linewidth}
      \centering
      \includegraphics[width=\linewidth]{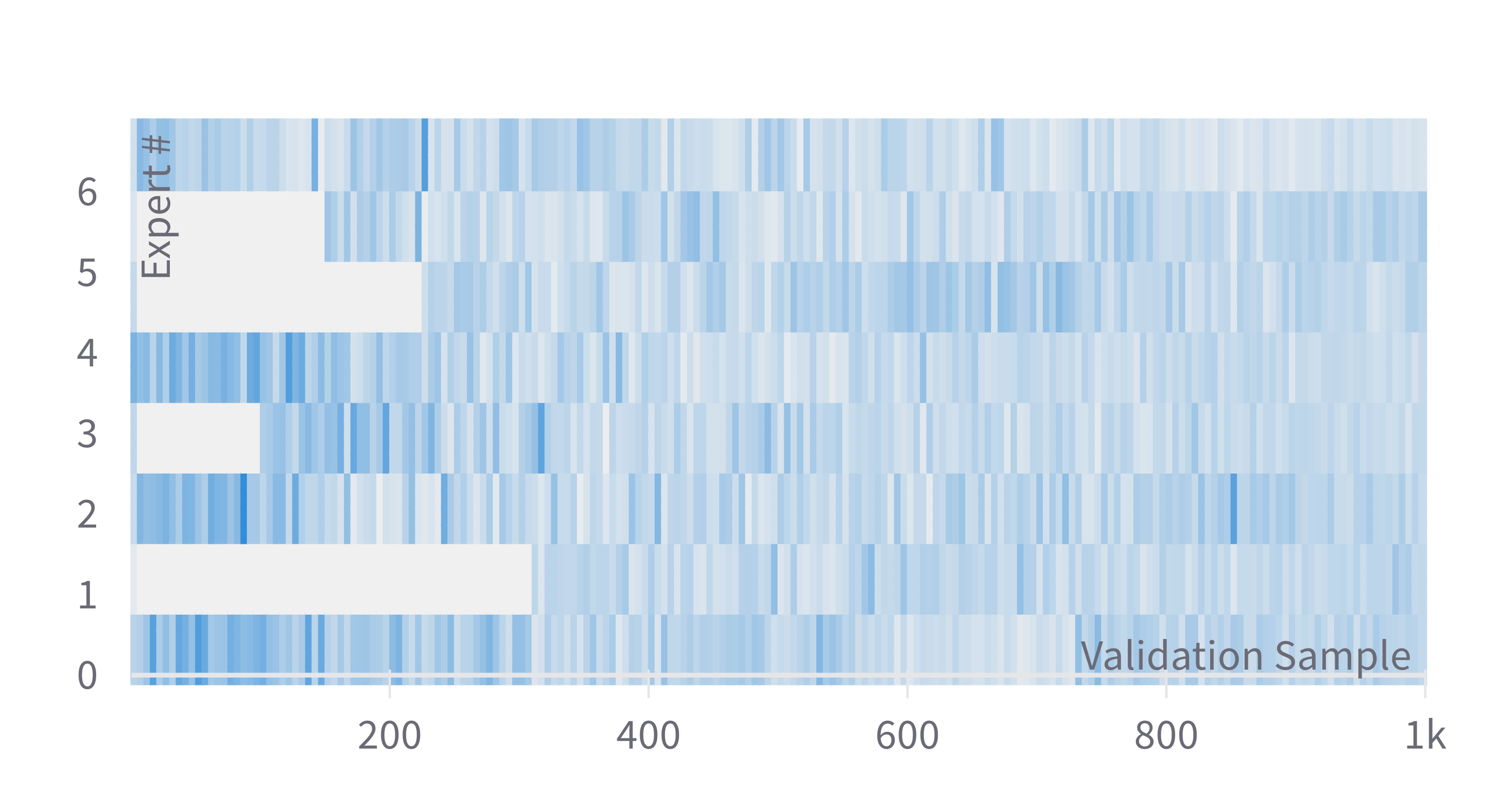}
      \caption{Adversarial training, entropy loss}
    \end{subfigure}
\caption*{BlockMoE replacing the 1st BasicBlock at stage $conv5\_x$ in \texttt{ResNet-18}}
\end{subfigure}

\begin{subfigure}{\linewidth}
    \begin{subfigure}[t]{.246\linewidth}
      \centering
      \includegraphics[width=\linewidth]{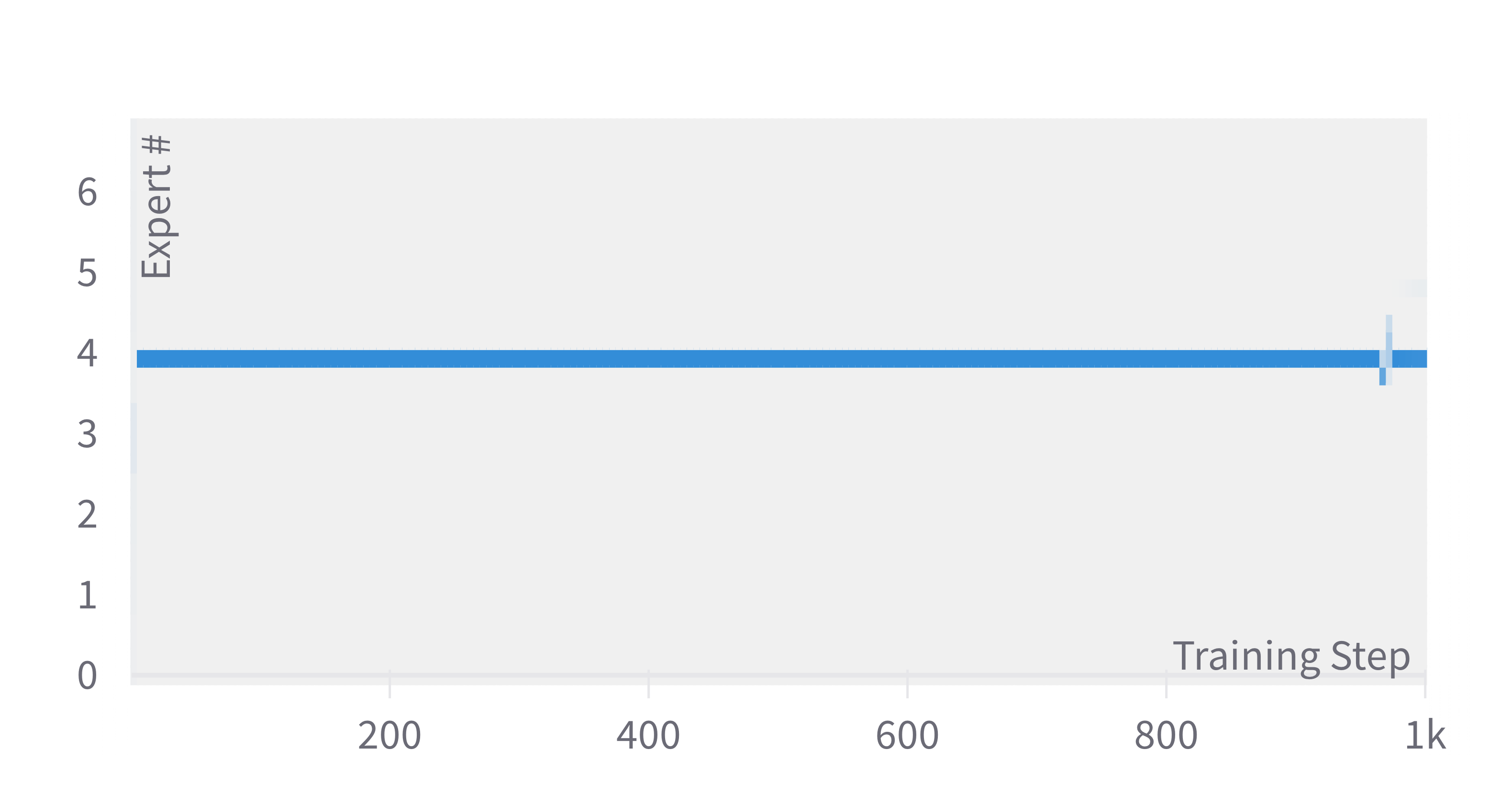}
      \caption{Normal training, switch loss.}
    \end{subfigure}
    \hfill
    \begin{subfigure}[t]{.246\linewidth}
      \centering
      \includegraphics[width=\linewidth]{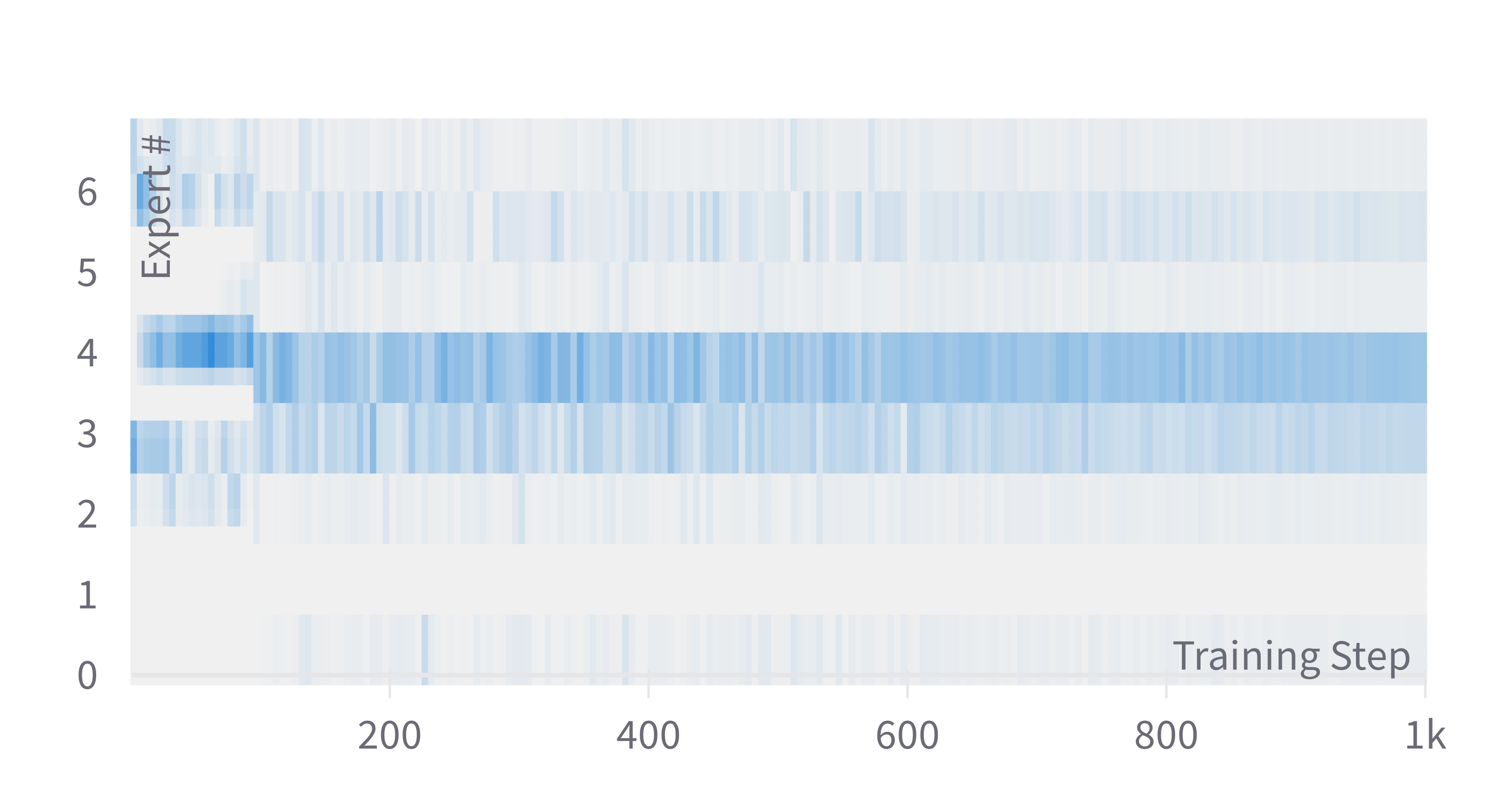}
      \caption{Normal training, entropy loss}
    \end{subfigure}
    \hfill
    \begin{subfigure}[t]{.246\linewidth}
      \centering
      \includegraphics[width=\linewidth]{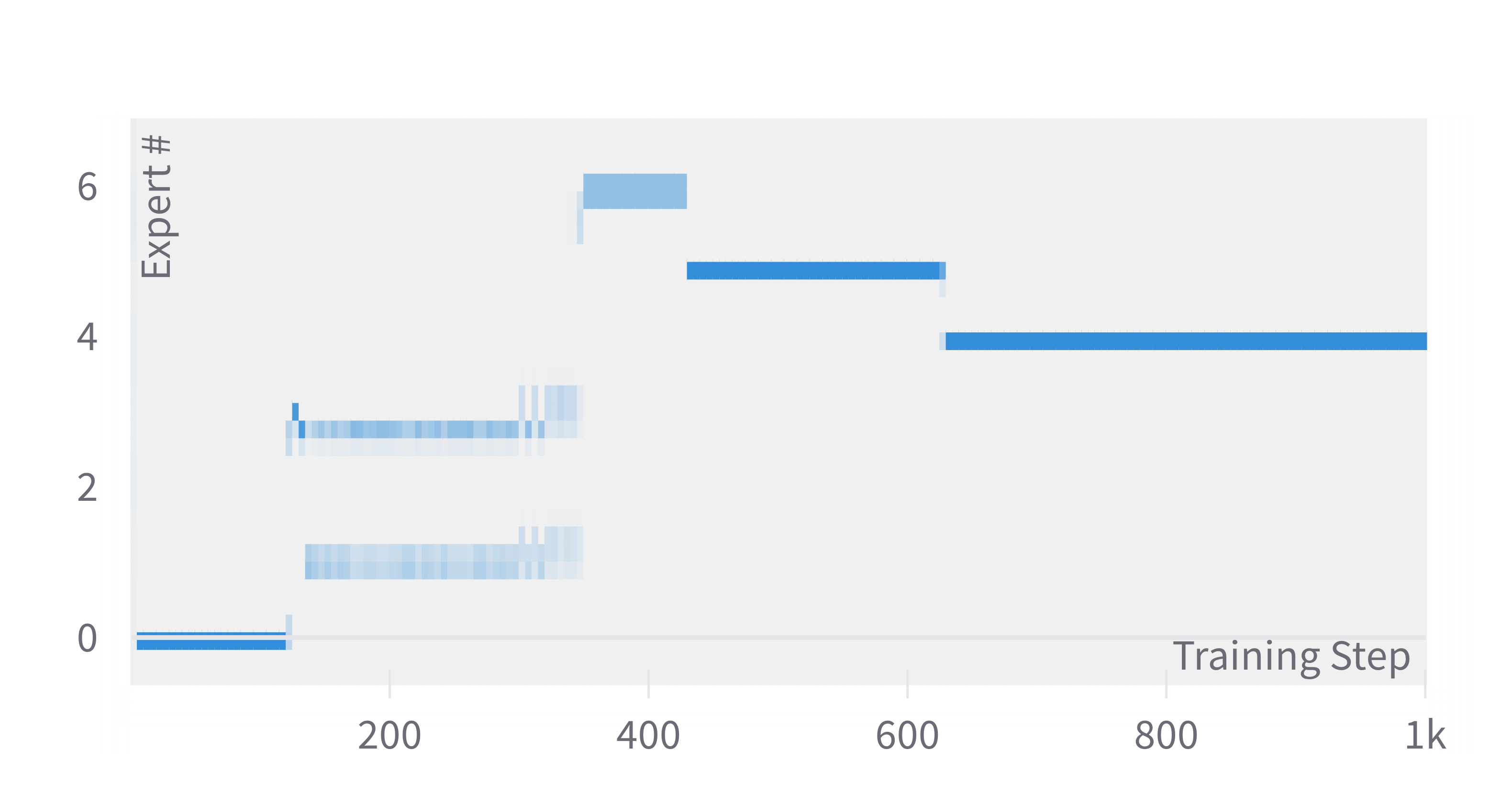}
      \caption{Adversarial training, switch loss}
    \end{subfigure}
    \hfill
    \begin{subfigure}[t]{.246\linewidth}
      \centering
      \includegraphics[width=\linewidth]{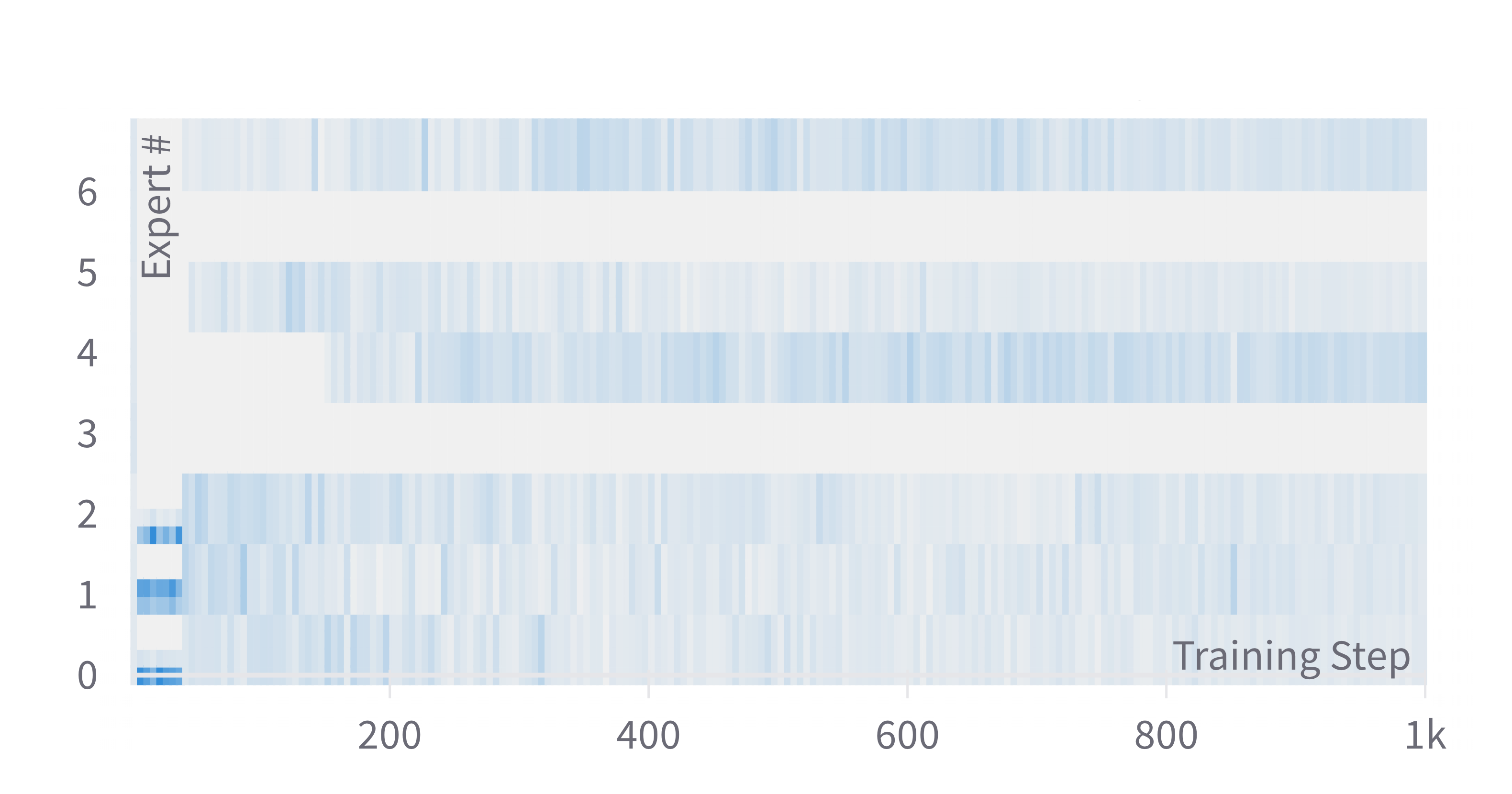}
      \caption{Adversarial training, entropy loss}
    \end{subfigure}
\caption*{BlockMoE replacing the 2nd BasicBlock at stage $conv5\_x$ in \texttt{ResNet-18}}
\end{subfigure}
\caption{Distribution of test inputs to individual experts during training for a \texttt{ResNet-18} with a BlockMoE layer with \textbf{eight experts}. Darker shades of blue correspond to more inputs being routed to this expert. The switch auxiliary loss is not able to prevent the dying expert phenomenon in the adversarial training case.}
\label{fig:routing-heatmaps}
\end{figure*}

\subsubsection{Routing Collapse}
\label{sec:vanishing_experts}

A common issue with MoE systems is experts that are not used at all, i.e., the so-called \textit{routing collapse}~\cite{shazeer2017outrageously} or  \textit{dying experts}~\cite{pavlitska2023sparsely}.
%From an optimization standpoint, this is considered a local minimum. 
Experts who receive more inputs early during training receive more informative gradient updates, are more likely to produce informative outputs quicker, and are, in turn, more likely to be chosen by the gating network.
%This self-perpetuating behavior prevents the MoEs from using the full amount of experts and, in turn, limits the number of parameters used at run time.
To mitigate it, we have employed two types of losses: switch and entropy losses.
Both losses encourage the uniform distribution of average importance scores and promote the gate to distribute inputs to all experts.

Figure~\ref{fig:routing-heatmaps} illustrates how inputs are routed to different experts in a BlockMoE layer with eight experts during training, under different conditions. The switch loss fails to prevent routing collapse. Extreme expert under-utilization can be observed in both normal and adversarial training.  In contrast, the entropy loss promotes a more balanced distribution across experts, especially for adversarial training, where multiple experts remain active. Finally, we could observe that the BlockMoE performs better when placed in the second residual block (deeper in the network), particularly under entropy loss. This placement allows for more meaningful routing and stable expert specialization. These findings are consistent with our previous experiments.

% For the standard neural network training, both losses are equally capable of achieving a close to uniform distribution of inputs to experts at the end of the training.
% However, the switch loss uses only a single expert for a certain number of steps before there is a sudden change to the usage of all experts.
% The entropy loss uses almost all experts from the start and gradually picks up the remaining ones early.
% For PGD adversarial training, the entropy loss exhibits essentially the same behavior, but the switch loss model fails to use all experts throughout the training.
% Even though some experts are selected for a part of the training, the selection process collapses to a single expert multiple times during the training and finalizes on expert 5.
% In this case, the model cannot use its additional parameters, making the whole MoE layer useless.

% For the second residual block in the final layer, both losses fail to encourage the usage of all experts, even though prior work suggests that MoEs are best placed as deep into the architecture as possible.
% Rajbhandari et al.~\cite{rajbhandari2022deepspeed} go as far as to use fewer experts in the first layers of the model and double the number of experts in the deeper layers.
% In this setting, tokens are distributed equally to experts in the earlier layer, while in the switch loss setting, only a single expert is used in the latest MoE layer.

\begin{figure*}[t]
\centering
\begin{subfigure}{\linewidth}
    \begin{subfigure}{0.246\linewidth}
      \centering
      \includegraphics[width=\linewidth]{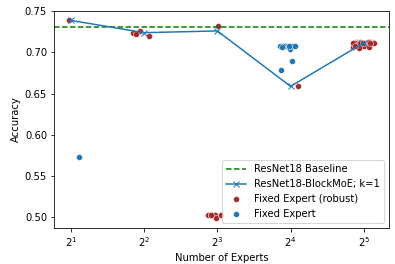}
      \caption{Clean data, switch loss}
    \end{subfigure}
    \hfill
    \begin{subfigure}{0.246\linewidth}
      \centering
      \includegraphics[width=\linewidth]{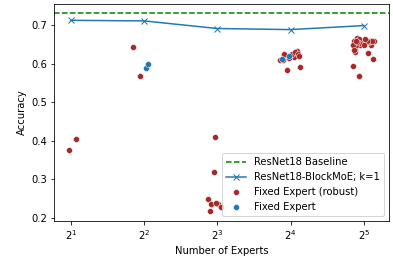}
      \caption{Clean data, entropy loss}
    \end{subfigure}
    \hfill
    \begin{subfigure}{0.246\linewidth}
      \centering
      \includegraphics[width=\linewidth]{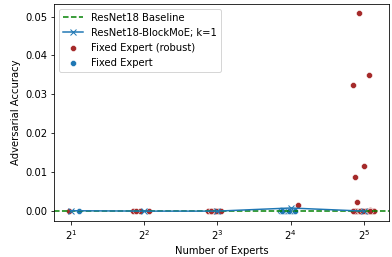}
      \caption{Under attack, switch loss}
    \end{subfigure}
    \hfill
    \begin{subfigure}{0.246\linewidth}
      \centering
      \includegraphics[width=\linewidth]{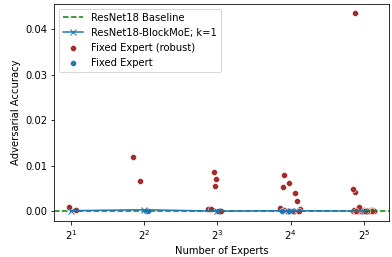}
      \caption{Under attack, entropy loss}
    \end{subfigure}
    \caption*{\textbf{Normal} training}
\end{subfigure}

\begin{subfigure}{\linewidth}
    \begin{subfigure}{0.246\linewidth}
      \centering
      \includegraphics[width=\linewidth]{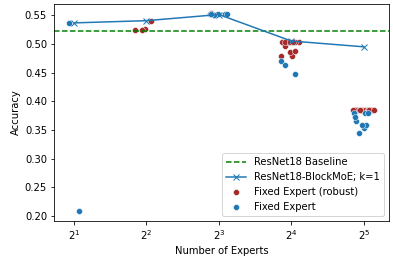}
      \caption{Clean data, switch loss}
    \end{subfigure}
    \hfill
    \begin{subfigure}{0.246\linewidth}
      \centering
      \includegraphics[width=\linewidth]{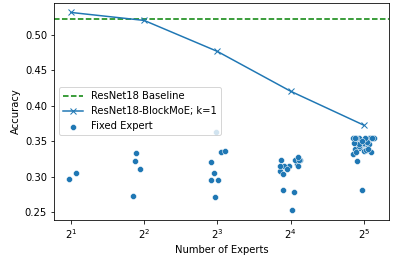}
      \caption{Clean data, entropy loss}
    \end{subfigure}
    \hfill
    \begin{subfigure}{0.246\linewidth}
      \centering
      \includegraphics[width=\linewidth]{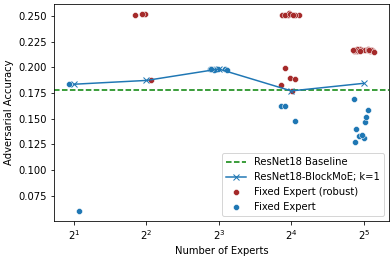}
      \caption{Under attack, switch loss}
    \end{subfigure}
    \hfill
    \begin{subfigure}{0.246\linewidth}
      \centering
      \includegraphics[width=\linewidth]{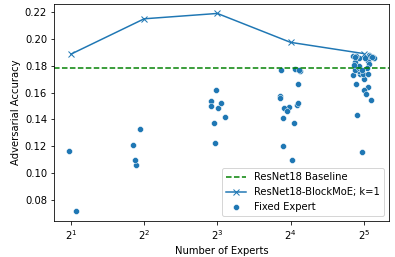}
      \caption{Under attack, entropy loss}
    \end{subfigure}
    \caption*{\textbf{Adversarial} training}
\end{subfigure}
\caption{Accuracy of the baseline (green), a model with an MoE layer (blue line), and models with a single fixed expert (dots) for normal and adversarially-trained \texttt{ResNet-18} models with BlockMoE layers for a varying number of experts. Experts resulting in a higher accuracy are marked brown; experts resulting in a lower accuracy are marked blue. The x-positions of the dots for fixed experts are slightly scattered for better visibility. Robust experts lie above the blue solid line, indicating that they outperform the full MoE model.}
\label{fig:fixed-expert-plots}

\end{figure*}

\subsubsection{Disabling the Learned Gate}

We further aim to understand the importance of the learned selection mechanism by systematically shutting it off and investigating the performance of individual experts.

Figure~\ref{fig:fixed-expert-plots} compares the accuracy for different auxiliary losses against the baseline \texttt{ResNet-18} model.
%There are several conclusions to draw from these plots.
As discussed previously, none of the models using MoE layers can outperform the baseline in any significant way.
There is even a slight decay in accuracy when scaling up to eight experts; this is most certainly related to the fact that most of these models are located within the expert layers, influencing the training dynamics and possibly requiring more extended training and/or adjusted hyperparameters.
However, when looking at the accuracy of using individual experts on the whole dataset, we can see that switch loss results in at least a single expert performing as well as the entire MoE system.
This aligns with Figure~\ref{fig:routing-heatmaps}, showing that only one expert is used.
We observe that no individual experts can keep up with the entire system for the entropy loss.
As we expect experts to specialize in specific subsets of the data, the accuracy drop during evaluation on the "wrong" inputs that the expert would usually not see is expected.

% \begin{figure}
% \centering
%     \begin{tabular}{cc}
       
%     \end{tabular}

% % \begin{subfigure}{0.45\linewidth}
% %   \centering
% %   \includegraphics[width=\linewidth]{img/classification/cifar_robust_switch_natural_fixed_expert_plot.png}
% %   \caption{Evaluation on clean data, importance auxiliary loss}
% % \end{subfigure}%
% % \begin{subfigure}{0.45\linewidth}
% %   \centering
% %   \includegraphics[width=\linewidth]{img/classification/cifar_robust_entropy_natural_fixed_expert_plot.png}
% %   \caption{Evaluation on clean data, KL-divergence auxiliary loss}
% % \end{subfigure}

% % \begin{subfigure}{0.45\linewidth}
% %   \centering
% %   \includegraphics[width=\linewidth]{img/classification/cifar_robust_switch_adv_fixed_expert_plot.png}
% %   \caption{Evaluation on attacked data, importance auxiliary loss}
% % \end{subfigure}%
% % \begin{subfigure}{0.45\linewidth}
% %   \centering
% %   \includegraphics[width=\linewidth]{img/classification/cifar_robust_entropy_adv_fixed_expert_plot.png}
% %   \caption{Evaluation on attacked data, KL-divergence auxiliary loss.}
% % \end{subfigure}

% \caption{Accuracy ofBlockMoE models trained with PGD-7 AT.  A single expert is fixed for the whole evaluation, i.e., replacing the MoE module with one of the experts. Experts with increased adversarial robustness are marked blue, and experts with lower adversarial accuracy are marked brown. The x-positions are slightly scattered for better visibility.}

% \label{fig:cifar_robust_fixed_expert_plot_auxiliary_losses}

% \end{figure}

Figure~\ref{fig:fixed-expert-plots} shows the resulting accuracies for individual experts and the whole expert system when trained using the PGD-7 adversarial training and evaluated on natural images or against a PGD adversary. % for the switch and entropy losses. 
%These plots reveal some other intriguing properties about adversarially trained MoE models.
In the case of entropy loss, individual experts perform roughly the same independent of the total number of experts when evaluated on natural images. When assessed on PGD-20 attacked inputs, the performance of individual experts is correlated with the total number of experts in that model.

In the case of the switch loss, an interesting phenomenon arises at the level of individual experts.
For unattacked images, using the learned gating mechanism to decide which expert should be activated depending on the current input is favorable. Only for the $32$ expert model is there a clear performance improvement of the learned gating compared to using any single expert. The MoE thus outperforms every single expert.
However, when the models are evaluated against an adversary, most models' gating mechanisms do not achieve the highest accuracy.

On the contrary, a group of significantly more robust experts emerges for three out of the five architecture versions.
With roughly $25\%$ adversarial accuracy for $4$ and $16$ experts and roughly $22\%$ for $32$ total experts, these groups of robust experts outperform the baseline by up to $7.5pp$.
These results indicate that sparse MoEs can be used as a proxy to train robust classification models by finding robust subpaths in networks with dynamic routing.

A possible explanation for the emergence of robust experts under switch loss is its tendency to route inputs through a smaller subset of experts, as shown in the test sample distributions (see Figure~\ref{fig:routing-heatmaps}). This concentrated routing may lead to more focused and consistent training for the selected experts, allowing them to specialize more deeply on frequently encountered patterns, including adversarial perturbations. As noted earlier, this focused specialization does not necessarily improve clean accuracy, but may help certain experts learn more robust internal representations.

\section{Conclusion}
Our work explored the potential of sparse MoE layers to improve adversarial robustness in CNNs. We applied MoEs at the level of residual blocks and convolutional layers and evaluated their effect under standard and adversarial training on \texttt{CIFAR-100}. We found that adversarially trained models with a single MoE layer placed in the deeper network stages achieved improved robustness to PGD and AutoPGD attacks, without sacrificing clean accuracy.

Comparing MoE layers at the level of residual blocks and convolutional layers, we could observe that BlockMoE layers achieve higher robustness, likely because coarser-grained modules such as residual blocks can learn more semantically distinct and independent representations, leading to more effective specialization under adversarial training.

Our analysis of routing dynamics has revealed that the commonly used switch loss often leads to routing collapse, with most inputs routed to a single expert, whereas the entropy loss more effectively encourages balanced expert utilization. Interestingly, despite this imbalance, we observed that some fixed experts trained with switch loss exhibited higher robustness than the full-gated MoE, suggesting that concentrated training on a few experts can inadvertently produce robust sub-networks. At the same time, we found that increasing the number of experts tends to reduce individual specialization, pointing to a trade-off between routing granularity and effective expert differentiation. These findings suggest that while entropy loss promotes diversity, the robustness benefits under switch loss may arise from focused adversarial attack on a limited set of experts.

While these results demonstrate the potential of MoEs in improving adversarial robustness, several limitations remain. Robustness gains were largely confined to adversarially trained settings. Under normal training, sparse MoEs offered little to no improvement. The fact that some individual experts performed better in isolation than the full MoE, as mentioned above, suggests that the gating may not consistently exploit the most robust computation paths.

Our findings motivate several future directions. Robustness-aware gating strategies could be developed to preferentially route inputs through resilient experts. Insights from the lottery ticket hypothesis~\cite{frankle2018lottery,fu2021drawing} could guide the identification and refinement of robust subnetworks within MoEs. Additionally, incorporating adversarial signals into the routing objective or dynamically adjusting expert selection at inference time could further improve performance. Together, these directions suggest that sparse MoEs are a promising architectural tool for adversarially robust deep learning.

\newpage
\clearpage
\section*{Acknowledgment}

This work was supported by funding from the Topic Engineering Secure Systems of the Helmholtz Association (HGF) and by KASTEL Security Research Labs (46.23.03).

{
    \small
    \bibliographystyle{ieeenat_fullname}
    \bibliography{references}
}

\end{document}